\documentclass[journal]{IEEEtran}

\usepackage{pkgs/algorithm}
\usepackage{pkgs/algorithmic}

\usepackage[colorlinks=true]{hyperref}
\usepackage[font=small]{caption}

\usepackage{graphics} 
\usepackage{amsmath} 
\usepackage{mathtools} 
\usepackage[utf8]{inputenc} 
\usepackage[english]{babel}

\usepackage{helvet}         
\usepackage{courier}        
\usepackage{type1cm}        
\usepackage{makeidx}         
\usepackage{graphicx}        
\usepackage{multicol}        
\usepackage[bottom]{footmisc}
\usepackage{subcaption} % used for subfigure side by sides
\usepackage{multirow}
\usepackage[table,xcdraw]{xcolor}

\usepackage{tikz}
\usetikzlibrary{positioning}
\usetikzlibrary{shapes,snakes}
\usepackage{bm} 
\usepackage{float} 
\usepackage{color} 
\usepackage{caption}
\usepackage{amsmath,amssymb} 
\usepackage{todonotes}

\usepackage{empheq} 

\usepackage[backend=bibtex,natbib=true,style=numeric-comp,sorting=none,firstinits=true,maxbibnames=99,url=false,doi=false]{biblatex}
\usepackage{upgreek,bm}
\usepackage{amsmath, amssymb, amsfonts}
\usepackage{todonotes}
\usepackage{float} 
\usepackage{xcolor} 
\usepackage{listings} 
\usepackage{pdflscape} 
\usepackage{booktabs} 
\usepackage{adjustbox} 
\usepackage{caption}

\usepackage{customCommands}

\newcommand\scalemath[2]{\scalebox{#1}{\mbox{\ensuremath{\displaystyle #2}}}}
\newcommand{\rev}[1]{#1}

\allowdisplaybreaks
 \renewcommand{\baselinestretch}{0.97}
\begin{document}
\title{Observability-Aware Intrinsic and Extrinsic Calibration of LiDAR-IMU Systems}

\author{Jiajun Lv$^{1,\dag}$, Xingxing Zuo$^{1,\dag}$, Kewei Hu$^1$, Jinhong Xu$^1$, Guoquan Huang$^2$, and Yong Liu$^{1,3}$
\thanks{$^1$The authors are with the Institute of Cyber-System and Control, Zhejiang University, Hangzhou, China.
Email: {\tt \{lvjiajun314, xingxingzuo, 21932133, xujinhong\}@zju.edu.cn, yongliu@iipc.zju.edu.cn}. (Yong Liu is the corresponding author.)} 
\thanks{$^2$The author is with the Department of Mechanical Engineering, University of Delaware, Newark, DE 19716, USA. Email: {\tt ghuang@udel.edu} }
\thanks{$^3$Huzhou Institute of Zhejiang University, Huzhou, 313000, China. }
\thanks{$^\dag$The co-first authors have equal contributions.}
}

% The paper headers
\markboth{IEEE Transactions on Robotics, VOL. XX, No. X, XXX 2022}%
{Lv \MakeLowercase{\textit{et al.}}: Observability-aware LiDAR-IMU Calibration}

\maketitle

\begin{abstract}
Accurate and reliable sensor calibration is essential to fuse  LiDAR and inertial measurements, which are usually available in robotic applications.
In this paper, we propose a novel LiDAR-IMU calibration method  within the continuous-time batch-optimization framework, 
where the intrinsics of both sensors and the spatial-temporal extrinsics between sensors are calibrated without using calibration infrastructure such as fiducial tags.
Compared to discrete-time approaches, the continuous-time formulation has natural advantages for fusing high rate measurements from LiDAR and IMU sensors. 
To improve efficiency and address degenerate motions, 
two observability-aware modules are leveraged:
(i) The information-theoretic data selection policy selects {\em only} the most informative segments for calibration during data collection, which significantly improves the calibration efficiency by processing only the selected informative segments. 
(ii) The observability-aware state update mechanism in nonlinear least-squares optimization  updates {\em only} the identifiable directions in the state space with truncated singular value decomposition (TSVD),
which enables accurate calibration results even under degenerate cases where informative data segments are not available.
The proposed LiDAR-IMU calibration approach has been validated extensively in both simulated and real-world experiments with different robot platforms,
demonstrating its high accuracy and repeatability in commonly-seen human-made environments.
We also open source our codebase to benefit the research community: {\url{https://github.com/APRIL-ZJU/OA-LICalib}}.
\end{abstract}


\IEEEpeerreviewmaketitle

\section{Introduction}

\IEEEPARstart{T}{he} LiDAR-IMU (LI) system has prevailed in recent years~\cite{zhang2014loam,shan2018lego,ye2019tightly,Zuo2020IROS,shan2020lio,lin2020decentralized,lv2021clins}, 
as an increasing number of robotic applications require accurate and robust LI-navigation solutions. Accurate LiDAR-IMU extrinsic calibration is a prerequisite of LI navigation.
Manually measuring the spatial extrinsic parameters (relative translation and rotation) between the LiDAR and IMU might be inaccurate or impractical,
demanding an easy-to-use, automatic calibration approach.
Besides the extrinsic calibration, calibrating the intrinsics of both sensors is equally important.
IMU intrinsics~\cite{li2014high,schneider2019observability} modeling scale imperfection and axis misalignment between the gyroscope and accelerometer, are recommended to calibrate offline~\cite{yang2020online}.
Similarly, LiDAR intrinsics~\cite{atanacio2011lidar,levinson2014unsupervised} such as each beam’s horizontal and vertical angles and range offset, have been shown to be better calibrated for accurate registration and mapping rather than using factory default. 
Therefore, in this paper, we perform both intrinsic and extrinsic calibration for LI systems.

Many calibration methods rely on infrastructure (e.g., fiducial targets, or turntables) to aid calibration~\cite{rehder2016extending, liu2019error, liu2019novel},
which however increases the technical barrier for the widespread deployment of LI systems. 
LI sensors used in practical applications may not be perfectly time synchronized, 
necessitating the temporal calibration to compute the time offset between the two sensors.
While discrete-time representation of states is most commonly used in calibration, 
its discretization introduces approximation when fusing asynchronous measurements and incurs a higher computational cost when processing high-rate data.
Building upon our prior work~\cite{lv2020targetless}, in this work, we develop a continuous-time LI calibration method 
that computes both intrinsic and extrinsic (spatial and temporal)  LI sensing parameters without using dedicated infrastructure.

It is well understood that (infrastructure-free) sensor calibration can be heavily affected by the environment and motion because of observability (e.g., see \cite{Zuo2020IROS,yang2020online,mirzaei20123d}). 
In order to make the calibration tool easy-to-use for an end user, 
it is desirable to enable our LI calibration method to automatically address observability issues.
To this end, we first perform observability-aware automatic selection of informative data for offline calibration; that is,  
we  select only the most informative trajectory segments (rather than an uninformative long-session trajectory with  weakly-excited motions). 
This is primarily due to the fact that 
if a calibration-data-collection trajectory is not fully excited, some calibration parameters may become unobservable and thus, cannot be computed or result in inconsistent results~\cite{yang2020online,mirzaei20123d}.
However, how to collect informative data for calibration is not trivial for non-expert users.
To alleviate this technology barrier, the proposed observability-aware data selection will automatically select informative trajectory segments,
which in turn significantly reduces the computational cost as less data is needed for accurate calibration.
On the other hand, if degenerate or weakly-excited motions inevitably occur --
for example, in the autonomous-driving case, straight-line translation without rotation, and single-axis rotation --
we also proposed to perform observability-aware state update during optimization so that no spurious information would influx into unobservable parameters.
As a result, the proposed observability-aware calibration method, termed as \textit{OA-LICalib}, 
is able to perform efficient and robust LI calibration  with challenging motions and scenes.

In particular, the main contributions of this work include:
\begin{itemize}
    \item The proposed OA-LICalib is among the first to perform full LI calibration that 
    is able to calibrate all sensing parameters including the spatial  (i.e., rigid-body transformation between the sensors) and temporal (i.e., time offset between the sensors) extrinsics, and the intrinsics of LiDAR and IMU.
    
    \item We propose the information-theoretic metrics to select informative trajectory segments for the LI calibration, 
    instead of naively processing all available data regardless its observability, which significantly reduces the computational cost.
    
    \item We develop the observability-aware state update during optimization by updating only the identifiable directions of state space under degenerate motions (e.g.,  planar motion and constant velocity).
    
    \item Extensive experiments in  simulations and real world are conducted to examine the feasibility and accuracy of the proposed OA-LICalib method, 
    whose codebase is also open sourced to broadly benefit the robotics community.
\end{itemize}

The remainder of this paper is organized as follows: 
After reviewing the literature in Section~\ref{sec:related_work}, we provide the background materials of continuous-time batch optimization and present the sensor models in Section~\ref{sec:sensor_model}. 
The problem formulation is given in Section~\ref{sec:method}, while the proposed observability-aware calibration method is presented in Section~\ref{sec:observability}. 
Simulated and real-world experiments are carried out in Section~\ref{sec:simu_experiment} and Section~\ref{sec:real_experiment}, respectively. 

Finally, the paper is concluded in Section~\ref{sec:conclusion}.
\section{Related Work}
\label{sec:related_work}

Sensor calibration has a rich literature.
Instead of providing a comprehensive literature review, in this section, we only review the closely related work of LI calibration.

\subsection{Spatial-temporal LI Calibration}

In order to calibrate the rigidly-connected LiDAR and IMU, 
Geiger et al.~\cite{geiger2013vision} proposed a motion-based calibration approach that performs the extrinsic calibration by hand-eye calibration~\cite{horaud1995hand}. However, their approach expects each sensor’s trajectory to be estimated independently and accurately, which is difficult for the consumer-grade IMU.  
Gentil et al.~\cite{le20183d} formulated the LI calibration as a factor graph optimization problem. Gaussian progress (GP) regression~\cite{williams2006gaussian} was adopted to up-sample IMU measurements for pose interpolation at every time instants of capturing LiDAR points. Although the IMU data is continuously modeled based on GP, only the states at some specific time instants are optimized in the discrete-time factor graph, which may reduce accuracy. 
Recently, Mishra et al.~\cite{mishra2021target} presented an EKF-based LI extrinsic calibration method integrated into the visual-inertial navigation framework OpenVINS~\cite{Geneva2020ICRA}.

Online extrinsic LI calibration approaches are also available. In~\cite{Zuo2019IROS,Zuo2020IROS}, LiDAR, IMU, and camera are tightly fused with a lightweight EKF,
where the extrinsic parameters between the sensors are estimated online.
Ye et al.~\cite{ye2019tightly} presented a tightly coupled LI odometry with online extrinsic parameters estimation, and adopted a linear motion model to eliminate motion distortion during the sweep. 
Qiu et al.~\cite{qiu2020real} developed an IMU-centric temporal offset and extrinsic rotation calibration method for the LI system based on the motion correlation.
Recently, Xu et al.~\cite{xu2021fast} proposed FAST-LIO2, which is a tightly-coupled LiDAR-inertial odometry with online extrinsic calibration in an iterated-EKF framework.
Note that online calibration often assumes reasonable initial values in order to guarantee convergence
and may have extra unobservable directions because of including calibration parameters into the state vector~\cite{rehder2016general,Zuo2020IROS}.

Continuous-time batch optimization with temporal basis functions is also widely studied in calibration problems. Furgale et al.~\cite{furgale2012continuous} detailed the derivation and realization for a full SLAM problem based on B-spline basis functions and evaluated the proposed framework within a camera-IMU calibration problem, which was further extended to support both temporal and spatial calibration~\cite{furgale2013unified}. 
Rehder et al.~\cite{rehder2014spatio} adopted a similar framework for calibrating the extrinsics between a camera-IMU system and a single-beam LiDAR in  two steps:
the camera-IMU system is first calibrated with a chessboard, and then the single-beam LiDAR is calibrated with respect to the camera-IMU system.

\subsection{Intrinsic Calibration}
Li et al.~\cite{li2014high} proposed an online state estimation framework using a multi-state-constraint Kalman filter (MSCKF) which simultaneously estimates extrinsic parameters of a visual-inertial system, time offset, and intrinsics of IMU and camera.
Rehder et al.~\cite{rehder2016extending} extended the previous work~\cite{furgale2013unified} to support the calibration of extrinsics and intrinsics of multiple IMUs in the visual-inertial system.
Yang et al.~\cite{yang2020online} investigated the necessity of online IMU intrinsic calibration in the visual-inertial system and performed observability analysis for different IMU intrinsic models. 

As the closest to our work, Liu et al.~\cite{liu2019error} proposed to calibrate a LI system's spatial extrinsics and intrinsics by discrete-time bundle adjustment optimization. A specially designed target composed of a cone and a cylinder is required to determine the intrinsics of 3D LiDAR and formulate the constraints involved with the rigid transformation between LiDAR and IMU. A high-precision 3-axis-adjustable turntable platform with accurate attitude readings is also necessary to calibrate the IMU intrinsics.
A close follow-up work~\cite{liu2019novel} simplifies the method~\cite{liu2019error} by removing the dedicated target requirement and providing the concept to calibrate with the multiple geometric features naturally existing in human-made environments. Point/sphere, line/cylinder, and plane features are extracted and leveraged for providing valid constraints. However, reliable and automatic geometric features extraction from real-world scenes can be error-prone and susceptible; the extra facility of a high-precision turntable platform remains necessary for IMU intrinsic calibration.

\subsection{Observability Awareness in Calibration}
It is known that we can assess the information content of trajectory segments for camera and IMU calibration~\cite{schneider2019observability}. 
As such, Maye et al.~\cite{maye2013self} proposed information-theoretic metrics to select informative data and discard redundant data for calibration. 
A truncated QR decomposition of the Fisher information matrix is employed to update estimates in only observable directions. 
Zhang et al.~\cite{zhang2016degeneracy} analyzed the geometric structures in optimized problem for point cloud registration and determined well-conditioned directions to solve them partially.
Schneider et al.~\cite{schneider2019observability} developed an online self-calibration method in discrete-time batch optimization for a visual-inertial system with the extended capability of calibrating IMU intrinsics. Multiple types of information metrics, including the trace, the determinant, and the maximal eigen value of the covariance matrix, are investigated for informative segment selection. The selected informative segments are fully excited with general motion, making the calibrated parameters observable and solvable.
Jiao et al.~\cite{jiao2021robust} designed a sliding window-based multi-LiDAR odometry system with the capability of online extrinsic calibration between multiple LiDARs, and the singular values of Hessian matrix are leveraged to examine the convergence of rotational extrinsic calibration.

\subsection{Extension of Our Previous Conference Publication~\cite{lv2020targetless}}
While this paper is evolved from our prior work on continuous-time extrinsic LI calibration of~\cite{lv2020targetless},
there are significant contributions differentiating this work from~\cite{lv2020targetless} and others.
To the best of our knowledge, the proposed OA-LICalib is among the first to fully calibrate all the sensing parameters of the LI sensors, 
including not only the spatial extrinsics as in~\cite{lv2020targetless}, 
but the temporal extrinsics (time offset) as well as the intrinsic parameters of the LiDAR and IMU.
As it is often required to fully excite the sensor platform during data collection in order for calibration results to converge,
which however might not be possible for some commonly-seen applications such as self-driving cars~\cite{Zuo2020IROS},
in this paper, we have proposed two observability-aware strategies to address this issue to make the proposed OA-LICalib method more easy-to-use for non-expert end users.
Specifically, (i) the information-theoretic based data selection policy selects the most informative trajectory segments for calibration, which eases the effort in calibration data collection and improves the calibration efficiency. 
(ii) The observability-aware state update scheme only updates the states (including calibration parameters) lying along the observable directions while preventing an influx of spurious information in the unobservable directions.

\section{Problem Formulation}
\label{sec:sensor_model}

\begin{figure}[t]
    \centering
    \includegraphics[width=1.0\columnwidth]{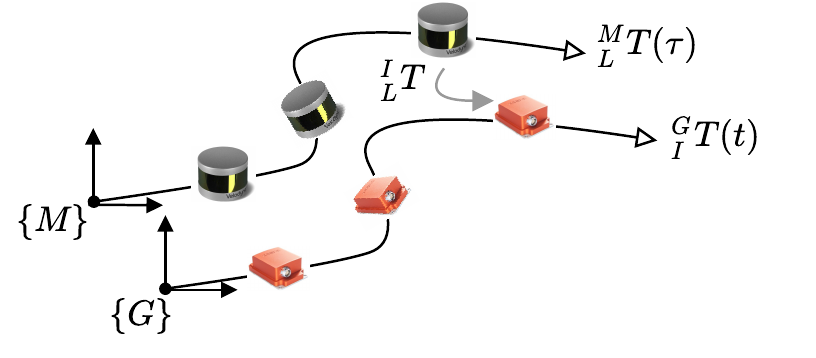}
    \caption{Illustration of different frames of reference used in the paper. 
    The trajectory of LiDAR can be represented with the IMU trajectory, the 6-DoF rigid transformation ${}^I_L\mathbf{T}$, and the time offset $t_c$ between LiDAR and IMU, by ${}^M_L\mathbf{T}(\tau)=
\left({}^G_I\mathbf{T}(\tau_0+t_c){}^I_L\mathbf{T}\right)^{\top}{}^G_I\mathbf{T}(\tau+t_c)\ {}^I_L\mathbf{T}$.}
    \label{fig:li_calib_frames}
\end{figure}

\begin{figure*}[t]
	\centering
	\includegraphics[width=1.0\textwidth]{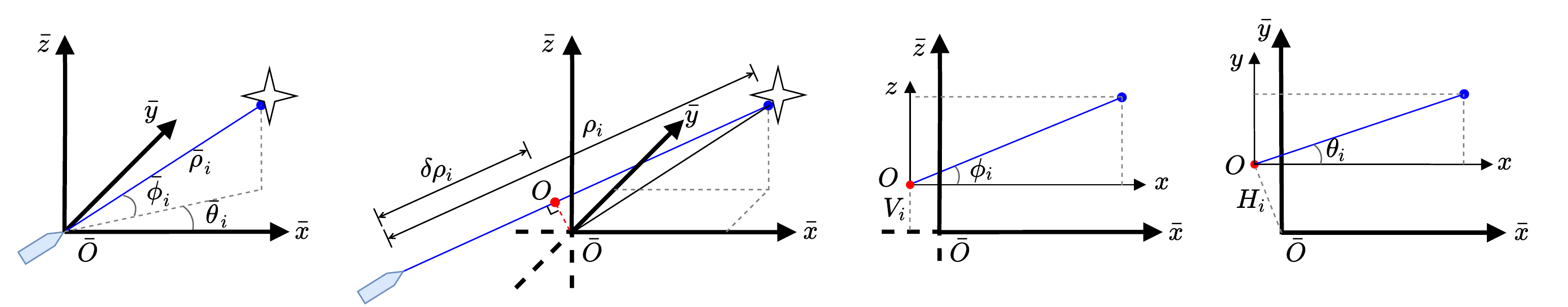}
	\caption{The intrinsics of an individual laser comprising a multi-beam 3D LiDAR. The left figure depicts the ideal model, while the three figures on the right illustrate the intrinsics arising from correction factors and offsets.} 
	\label{fig:lidar_intrinsic}
\end{figure*}

We first present our notations used throughout the paper.
We denote the 6-DoF rigid transformation by ${}^B_A\mat{T}\in \SE(3) \subset \R^{4\times4}$, which transforms the point ${}^A\vect{p}\in \R^3$ in frame $\{A\}$ into frame $\{B\}$. ${}^B_A\mat{T} = \begin{bmatrix} {}^B_A\mat{R} &  {}^B\vect{p}_A \\ \mathbf{0} & 1\end{bmatrix}$ consists of rotational part ${}^B_A\mat{R}\in \SO(3)$ and translational part ${}^B\vect{p}_A\in \R^3$. For simplicity, we omit the homogeneous conversion in the rigid transformation by $ {}^B\vect{p} = {}^B_A\mat{T}  {}^A\vect{p}$. 
${}^B_A\bar{q}$ is the quaternion of rotation corresponding to the rotation matrix ${}^B_A\mathbf{R}$. 
In our formulation, IMU reference frame $\{I\}$ is assumed to be rigidly connected to the LiDAR frame $\{L\}$. The map frame $\{M\}$ of LiDAR point cloud is determined by the first LiDAR frame $\{L_0\}$ when the calibration starts. Similarly, the global frame $\{G\}$ is determined as the first IMU frame $\{I_0\}$.
The spatial extrinsics, including the relative rotation and translation from the LiDAR frame to the IMU frame, are denoted by ${}^I_L\bar{q}$ and ${}^I\mathbf{p}_L$, respectively.
The temporal extrinsic parameter modeling the time offset between LiDAR and IMU is $t_c$. The LiDAR scan labeled with timestamp $\tau_L$ corresponds to IMU time instant $t_I = \tau_L + t_c$. Figure~\ref{fig:li_calib_frames} visualizes the different frames of reference used in this paper.

The extrinsic and intrinsic LI calibration considered in this paper aims to estimate the following states:
\begin{align}
	\mathcal{X} = \{ \bx_p, \bx_q, \bx_{Is}, \bx_I, \bx_L, {}^{I}_{L}\bar{q}, {}^{I}\mathbf{p}_{L}, t_{c}\}
\end{align}
which includes  the control points of B-splines $\{\bx_p, \bx_q\}$ to represent the continuous-time trajectory,
the IMU navigation states $\bx_{Is}$, 
the IMU intrinsics $\bx_I$, 
the LiDAR intrinsics $\bx_L$, 
and the LiDAR-IMU spatiotemporal extrinsic parameters $ \{{}^{I}_{L}\bar{q}, {}^{I}\mathbf{p}_{L}, t_{c}\} $.
Based on the LI sensor data collected during the calibration process, 
we formulate the following nonlinear least-squares (NLS) problem:
\begin{align}
  \label{eq:least-squares}
  \hat{\mathcal{X}} = \underset{\mathcal{X}}{\operatorname{argmin}} \;r, \;\;  r= r_I + r_L \text{,}
\end{align}
where  $r_I$ is the IMU measurement residual cost function which will be derived in detail in \eqref{eq:imuloss} 
and $r_L$ is the LiDAR measurement residual cost function as derived in \eqref{eq:lidarloss}.

\subsection{Continuous-time Trajectory Representation}
\label{sec:trajectory_represetation}

We employ B-spline to parameterize trajectory as it provides closed-form analytic derivatives~\cite{sommer2019efficient}, enabling the effortless fusion of high-frequency measurements for state estimation. B-spline also has the good property of being locally controllable, which means the update of a single control point only impacts certain consecutive segments of the spline~\cite{furgale2012continuous}. This trait yields a sparse system with a limited number of control points.
To parameterize the continuous-time 6-DoF trajectory, we leverage the split representation~\cite{haarbach2018survey} which represents the 3D translation and the 3D rotation with uniform B-splines separately. 
Both the translation and the rotation splines are parameterized with cumulative form. 
To be specific, the translation $\mathbf{p}(t)$ of $d$ degrees over time $t\in [t_i, t_{i+1})$ is controlled by the temporally uniformly distributed translational control points $\mathbf{p}_i$, $\mathbf{p}_{i+1}$, $\dots$, $\mathbf{p}_{i+d}$, and the matrix format could be expressed:
\begin{align}
\label{Eq:p_curve_cumu}
\mathbf{p}(t) = \mathbf{p}_i + \sum_{j=1}^{d} \mathbf{u}^{\top}{\Tilde{\mathbf{M}}}^{(d+1)}_{(j)}
\left(\mathbf{p}_{i+j}-\mathbf{p}_{i+j-1} \right)
\end{align}
where $\mathbf{u}^{\top} = \begin{bmatrix} 1 & u & \dots & u^d \end{bmatrix}$ and $u=(t-t_i)/(t_{i+1}-t_i)$. 
$\Tilde{\mathbf{M}}^{(4)}_{(j)}$ is the j-th column of the cumulative spline matrix $\Tilde{\mathbf{M}}^{(4)}$ which only depends on the corresponding degree of uniform B-spline. In this paper, cubic ($d=3$) B-spline is employed, thus the corresponding cumulative spline matrix is
\begin{align}
\Tilde{\mathbf{M}}^{(4)} =  \frac{1}{6}
\begin{bmatrix} 6 & 5 & 1 & 0 \\
0 & 3 & 3 & 0 \\
0 & -3 & 3 & 0 \\
0 & 1 & -2 & 1 \\
\end{bmatrix}
\text{.}
\end{align}
Besides, we adopt the cumulative representation of B-spline to parameterize the rotation on $\mathit{SO}(3)$ like~\cite{ovren2019trajectory,haarbach2018survey,sommer2020efficient}, and quaternions are served as the rotational control points of B-splines~\cite{kim1995general}: 
\begin{align}
\label{Eq:q_curve_cumu}
\bar{q}(t) = \bar{q}_i \otimes \prod_{j=1}^{d} exp\left(\mathbf{u}^{\top}{\Tilde{\mathbf{M}}}^{(4)}_{(j)}log(\bar{q}_{i+j-1}^{-1}\otimes\bar{q}_{i+j})\right)
\end{align}
where $\otimes$ denotes quaternion multiplication. 
$\bar{q}_i$ denotes the rotational control point. $exp\left(\cdot\right)$ is the operation that mapping Lie algebra elements to $\mathit{S}^3$, and $log\left(\cdot\right)$ is its inverse operation~\cite{kim1995general}. 

In this calibration system, the continuous 6-DOF poses of IMU ${}_I^{G}\bar{q}(t), {}^{G}{\mathbf{p}}_I(t)$ are represented in the global frame $\{G\}$ by splines in the format of \eqref{Eq:p_curve_cumu} and  \eqref{Eq:q_curve_cumu} with parameters of control points $\bx_p$ and $\bx_q$ for 3D translation and rotation, respectively. 
The derivatives of the splines with respect to time can be easily computed \cite{sommer2020efficient}, which lead to linear accelerations ${}^I\mathbf{a}(t)$ and angular velocities ${}^I\bomega(t)$ in the local IMU reference frame:
\begin{align}
\label{eq:spline_accel}
{}^I\mathbf{a}(t) &=
{}_I^{G}\mathbf{R}^{\top}(t) \left( {}^{G}\ddot{\mathbf{p}}_I(t) - {}^{G}\mathbf{g}\right) \\
\label{eq:spline_omega}
{}^I\bomega(t) &= {}_I^{G}\mathbf{R}^{\top}(t) {}_I^{G}\dot{\mathbf{R}}(t)
\end{align}
where $^G\mathbf{g} \in \mathbb{R}^3$ denotes the gravity vector in global frame. $^G\mathbf{g}$ has only two degrees of freedom since its norm is assumed to be a constant $\|{}^G\mathbf{g}\| \simeq 9.8$ in the system.

\subsection{IMU Model}
IMU comprises a 3-axis gyroscope and a 3-axis accelerometer. Inspired by~\cite{li2014high,rehder2016extending,yang2020online}, the IMU measurements are modeled as:
\begin{align} 
		{}^{\omega}\bomega &= \mathbf{S}_{\omega} \mathbf{M}_{\omega}\ {}^\omega_I\mathbf{R} {}^{I}\bomega(t)+\mathbf{A}_{\omega}{}^{I}\mathbf{a}(t)+ \mathbf{b}_{\omega}+\mathbf{n}_{\omega} \notag \\ 
		&= h_{I_\omega}(t,\bx_q, ^G\mathbf{g}, \mathbf{S}_{\omega}, \mathbf{M}_{\omega}, {}^\omega_I\mathbf{R}, \mathbf{A}_{\omega}, \mathbf{b}_{\omega}) 
		\label{eq:imu_w_model}
		\\
		{}^{a}\ba &=
		\mathbf{S}_{a} \mathbf{M}_{a}\ {}^a_I\mathbf{R} {}^{I}\mathbf{a}(t) +\mathbf{b}_a+\mathbf{n}_a \notag \\ 
		&= h_{I_a}(t, \bx_p, \bx_q, ^G\mathbf{g}, \mathbf{S}_{a}, \mathbf{M}_{a}, {}^a_I\mathbf{R}, \mathbf{b}_{a} ) 
		\label{eq:imu_a_model}
\end{align}
where $\bx_p$ and $\bx_q$ are the translational and rotational control points of B-splines, respectively. $\mathbf{b}_\omega, \mathbf{b}_a$ are the biases of gyroscope and accelerometer, which are assumed to be under Gaussian random walk. $\mathbf{n}_\omega$ and $\mathbf{n}_a$ are the zero-mean white Gaussian noise with covariance matrix ${\Sigma_\omega}$ and ${\Sigma_a}$, respectively, which can be obtained from IMU noise model without approximation or propagation.
$\mathbf{S}_{\omega/a}$ is a diagonal matrix modeling the scale imperfection as 
\begin{align}
    \mathbf{S}_{\omega} = 
    \begin{bmatrix} 
        S_{w1} & 0 & 0  \\
        0 & S_{w2} & 0  \\
        0 & 0 & S_{w3} \\
    \end{bmatrix}\text{,}\ 
    \mathbf{S}_{a} = 
    \begin{bmatrix} 
        S_{a1} & 0 & 0  \\
        0 & S_{a2} & 0  \\
        0 & 0 & S_{a3} \\
    \end{bmatrix}\text{.}
\end{align}
$\mathbf{M}_{\omega/a}$ accounting for axis misalignment is an upper-triangular matrix with diagonal elements as identity
\begin{align}
    \mathbf{M}_{\omega} = 
    \begin{bmatrix} 
        1 & M_{w1} & M_{w2}  \\
        0 & 1 & M_{w3}  \\
        0 & 0 & 1 \\
    \end{bmatrix}\text{,}\ 
    \mathbf{M}_{a} = 
    \begin{bmatrix} 
        1 & M_{a1} & M_{a2}  \\
        0 & 1 & M_{a3}  \\
        0 & 0 & 1 \\
    \end{bmatrix}\text{.}
\end{align}
$\bA_\omega$ is a full $3\times3$ matrix accounting for the acceleration dependence (g-sensitivity) of the measurements.

Since the gyroscope $\{\omega\}$ and accelerometer $\{a\}$ inside the IMU are individual sensors, there is a misalignment between them and the base IMU frame $\{I\}$. ${}^{\omega/a}_I\mathbf{R}$ compensates for the rotational misalignment, and the translational misalignment is omitted for calibration since it is close to zero for single-chip MEMS sensors~\cite{zachariah2010joint,schneider2019observability, yang2020online}. We can only calibrate ${}^{\omega}_I\mathbf{R}$ or ${}^{a}_I\mathbf{R}$ since the base IMU frame $\{I\}$ is aligned with either $\{\omega\}$ or $\{a\}$, and calibrating both makes the system unobservable~\cite{yang2020online}.
In this paper, we choose to calibrate ${}^{\omega}_I\mathbf{R}$ only. Additionally, we ignore the calibration of gravity sensitivity in analogy to~\cite{yang2020online}.

Usually, IMU measurements are given at discrete time instants. The measurements at timestamp $t_k$ are denoted as ${}^{\omega_k}\bomega_m, {}^{a_k}\ba_m$. Since the trajectory is formulated in continuous time, we can readily get ${}^I\mathbf{a}(t_k), {}^I\bomega(t_k)$ from \eqref{eq:spline_accel} and \eqref{eq:spline_omega} by computing differentials of the trajectory. The IMU induces cost functions, $r_I$, related to estimated variables including the control points of the trajectory $\{\bx_p, \bx_q\}$, the IMU navigation states $\bx_{Is} = \{^G\mathbf{g}, \mathbf{b}_{\omega}, \mathbf{b}_{a}\}$, and the IMU intrinsics $\bx_{I} = \{\mathbf{S}_{\omega}, \mathbf{M}_{\omega}, {}^\omega_I\mathbf{R}, \mathbf{S}_{a}, \mathbf{M}_{a}\}$ are formulated as:
\begin{align}
	&r_I = r_{\omega} + r_a\text{,}  \quad \text{where} \label{eq:imuloss}\\
	&r_{\omega} = \sum_{k} \frac{1}{2}\|{}^{\omega_k}\bomega_m \!-\!  h_{I_\omega}(t_k, \bx_q, \bx_I, \bx_{Is}) \|^2_{\Sigma_\omega} \text{,}  \label{eq:gyro_loss} \\ 
	&r_a = \frac{1}{2} \| {}^{a_k}\ba_m \!-\! h_{I_a}(t_k, \bx_p, \bx_q,  \bx_I, \bx_{Is} ) \|^2_{\Sigma_a}\text{,}\label{eq:accel_loss}
\end{align}
$\|\mathbf{e} \|^2_\Sigma = \mathbf{e}^\top \Sigma^{-1} \mathbf{e}$ denotes the squared energy norm weighted by the inverse of the covariance matrix $\Sigma$, and $r_{\omega}, r_a $ are residuals of angular velocity and linear acceleration according to \eqref{eq:imu_w_model} and \eqref{eq:imu_a_model}, respectively.

\subsection{3D LiDAR Model} \label{sec:lidar model}
 A mechanical 3D LiDAR with multiple laser beams~\cite{glennie2010static} measures the ranges given by individual laser heads pointing to different elevation angles. A LiDAR scan is captured when the rigidly connected laser heads rotate around a mechanical central axis. A laser head $i$ with elevation angle $\phi_{i}$ at LiDAR time instant $\tau_k$ (with azimuth angle $\theta_{ik}$) gets a range measurement  $\rho_{ik}$, then the LiDAR point measurement transferred from the spherical coordinate system can be denoted by:
 \begin{align}
 {}^{L_{k}}\bp_{ik} = \begin{bmatrix}  {}^{L_{k}} x_{ik} \\  {}^{L_{k}} y_{ik} \\ {}^{L_{k}} z_{ik}  \end{bmatrix} =  \begin{bmatrix} \rho_{ik} \cos\phi_i\cos\theta_{ik} \\ \rho_{ik} \cos\phi_i\sin\theta_{ik} \\ \rho_{ik}\sin\phi_{i} \end{bmatrix}\text{.}
 \end{align}
However, the above ideal model does not hold in practical manufacture since spatial offsets exist between laser heads, and non-negligible errors exist in both the range measurement and the azimuth angle $\theta_{ik}$. As shown in Fig.~\ref{fig:lidar_intrinsic}, the intrinsics of an individual laser head $i$ consist of: 
\begin{itemize}
    \item Elevation angle correction factor $\delta \phi_i$ and azimuth angle correction factor $\delta \theta_i$.
    \item Vertical spatial offset $V_i$ and  horizontal spatial offset $H_i$.
    \item Scale factor of range measurement $s_i$ and  range measurement offset $ \delta \rho_{i}$.
\end{itemize}
Given the intrinsics, the zero-mean Gaussian measurement noise $n_{\rho,ik}$ over the range measurement, and defining the following factors:
\begin{equation}
\left\{
\begin{array}{c}
     \bar{\phi}_i = \phi_i + \delta \phi_i \\
     \bar{\theta}_{ik} = \theta_{ik} + \delta \theta_i \\
     \bar{\rho}_{ik} = s_i \rho_{ik} + \delta \rho_{i} + n_{\rho,ik}
\end{array}
\right.
\end{equation}
The position of LiDAR point measurement is given by:
\begin{align}
	{}^{L_{k}}\bp_{ik} = \begin{bmatrix} \bar{\rho}_{ik} \cos\bar{\phi}_i\cos\bar{\theta}_{ik} + H_i \sin\bar{\theta}_{ik} \\ \bar{\rho}_{ik} \cos\bar{\phi}_i\sin\bar{\theta}_{ik}  + H_i \cos\bar{\theta}_{ik} \\ \bar{\rho}_{ik}\sin\bar{\phi}_{i} + V_i \end{bmatrix}\text{.}
\end{align}
The intrinsics of a 3D LiDAR with $l$ beams are denoted by 
\begin{align}
\bx_L = \{\delta \phi_i, \delta \theta_i, V_i, H_i, s_i,  \delta \rho_{i}\}_{|_{i=0,1,\cdots,l-1}}\text{.}
\end{align}

For a LiDAR point measurement ${}^{L_k}\bp_{ik}$ with an associated 3D plane with closest point parameterization~\cite{geneva2018lips}, ${}^{M}\bpi_j = {}^{M}d_{\pi,j}{}^{M}\bn_{\pi,j}$, where ${}^{M}d_{\pi,j}$ and ${}^{M}\bn_{\pi,j}$ denote the distance of the plane to origin and unit normal vector expressed in map frame, respectively.
The point to plane distance is given by:
\begin{align}
	{}^{M}\bp_{ik} &= {}^{M}_{L_{k}}\mathbf{R}(\tau_k) {}^{L_{k}}\bp_{ik} + {}^{M}\mathbf{p}_{L_{k}}(\tau_k) \\
	\bz_{ijk} &= {}^{M}\bn^\top_{\pi,j} {}^{M}\bp_{ik} + {}^{M}d_{\pi,j} \notag\\
	&=h_L(\tau_k, \bx_p, \bx_q, \bx_L, {}^{I}_{L}\bar{q}, {}^{I}\mathbf{p}_{L}, t_{c})
	\text{.}
\label{eq:point2planedis}
\end{align}
The distance measurement above is related to variables including the control points of trajectory $\{\bx_p, \bx_q\}$, LiDAR intrinsics $\bx_L$, and spatial-temporal parameters $ \{{}^{I}_{L}\bar{q}, {}^{I}\mathbf{p}_{L}, t_{c}\} $. 
The 3D plane is extracted from LiDAR point cloud map, detailed in Sec.\ref{sec:data_association}, and assumed free of noise when computing $\bz_{ijk}$. Thus, ${\Sigma_{z_{ijk}}}$, the covariance of the distance measurement $\bz_{ijk}$, can be easily propagated from the Gaussian noise $n_{\rho,ik}$ by first-order approximation.
The relative pose $\{{}^{M}_{L_{k}} \mathbf{R}(\tau_k), {}^{M}\mathbf{p}_{L_{k}}(\tau_k)\}$ in \eqref{eq:point2planedis} is involved with the following equations from parameterized IMU continuous-time trajectory:
\begin{align}
    {}^M_{L_k}\mathbf{T}(\tau_k) =
    \left( {}^G_I\mathbf{T}(\tau_0+t_c){}^I_L\mathbf{T}\right)^{\top}{}^G_I\mathbf{T}(\tau_k+t_c)\ {}^I_L\mathbf{T}
    \text{.}
\end{align}
It is easy to find that the time offset $t_{c}$ between LiDAR and IMU is also involved in the process of fetching pose ${}^{M}_{L_{k}} \mathbf{T}(\tau_k)$ from continuous-time trajectory parameterized in the IMU time axis.
In practice, the LiDAR point measurement at timestamp $\tau_k$ is denoted as $\bz_{m,ijk}$. By minimizing the point to plane distance, we have the following cost functions from LiDAR points associated with 3D planes:
\begin{align}
	\label{eq:lidarloss}
	r_L \!=\! \scalemath{0.9}{\frac{1}{2} \sum_{i}  \sum_{j}\sum_{k} \|\bz_{m,ijk} - h_L(\tau_k, \bx_p, \bx_q, \bx_L, {}^{I}_{L}\bar{q}, {}^{I}\mathbf{p}_{L}, t_{c}) \|^2_{\Sigma_{z_{m,ijk}}}}
\end{align}
where $i,j,k$ denote the indexes of LiDAR beams, 3D planes and points in a scanned ring by a laser beam, respectively. 

\begin{figure*}[t]
    \centering
    \includegraphics[width=1.0\textwidth]{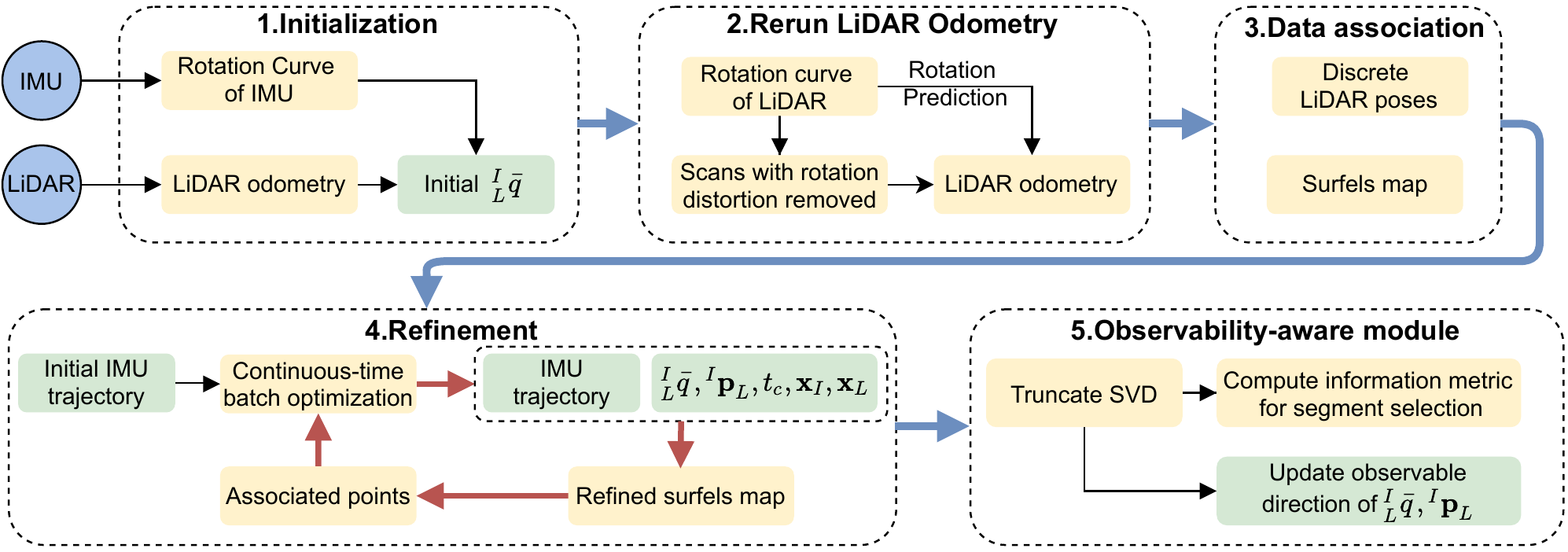}
    \caption{The pipeline of proposed LiDAR-IMU calibration method, which allows leveraging all the raw measurements from IMU and LiDAR sensor in a continuous-time batch optimization framework.  Details are provided in Section~\ref{sec:method} and Section~\ref{sec:observability}. }
    \label{fig:pipline}
\end{figure*}

\section{Continuous-time Full LI Calibration}
\label{sec:method}

In this section, we present in detail the proposed continuous-time full LI calibration method.
Before diving into details, we first provide an overview of the overall architecture as shown in Fig.~\ref{fig:pipline}.
Specifically, we first initialize the control points of rotation B-splines with the IMU angular velocities, 
and the extrinsic rotation ${}^I_L\bar{q}$ by aligning the IMU rotations with the LiDAR odometry (rotations) computed with NDT registration~\cite{magnusson2009three}.
By reducing the rotational distortion effect in LiDAR scans, we re-compute the LiDAR odometry to improve its accuracy.
We then initialize the LiDAR surfels  by fusing LiDAR scans with the computed LiDAR odometry and find the point-to-surfel correspondences.
With the LI measurements, we perform batch optimization to estimate the considered states including B-splines control points and spatial-temporal extrinsics,
whose optimal estimates are used to refine the surfels map and point-to-plane data association.

\subsection{State Initialization}
\label{sec:rotation_init}

Initial guesses of the calibrated parameters are needed for the iterative non-linear least-squares optimization. For initialization, we set IMU navigation states as $\hat{\bx}_{{Is}_0} = \{\|^G \mathbf{g}\|\be_3, \mathbf{0}_3, \mathbf{0}_3\}$ and IMU intrinsics as $\hat{\bx}_{I_0} = \{\bI_3, \bI_3, \bI_3, \bI_3, \bI_3\}$, where $\be_3 = [ 0, 0, 1]^\top$, $\bI_3$ and $\mathbf{0}_3$ represent the $3 \times 3$ identity matrix and zero matrix, respectively.
As for the initial values of LiDAR intrinsics, $\hat{\bx}_{L_0} $ is initialized as $ \{0, 0, 0, 0, 1,  0\}_i{|_{i=1,\cdots,l}}$. 
The time offset $t_{c}$ between LiDAR and IMU is initialized as zero, which means the timestamp $t_k$ of IMU is equal to the timestamp $\tau_k$ of LiDAR at the beginning.

The initialization of control points and extrinsic parameters are described in detail below. Prior to detailing the initialization process, we clarify the inputs for initialization including discrete poses at about 10 Hz from LiDAR odometry, angular velocities and linear accelerates at typically 400 Hz from IMU.
We adopt NDT registration to implement simple scan-to-map LiDAR odometry, sufficient for sequences with a few seconds, and build a global map consists of keyframed LiDAR scans determined according to the pose changes.

\subsubsection{Initialize Control Points of Continuous-time Trajectory}
\label{sec:initialization of control points}

Before detailing the concrete initialization method, we emphasize that the spline trajectory is employed to represent the IMU trajectory rather than LiDAR's. If we model the continuous-time trajectory in LiDAR sensor, the predicted linear acceleration and angular velocity in IMU frame need to be transformed from LiDAR frame as
\begin{align}
    {}^I\vect{a}(t) &= {}^I_L\mat{R} (
    {}^M_L\mat{R}(t)^{\top}\left({}^M\vect{a}(t) - {}^M\vect{g}\right) \notag \\
    & + \lfloor{{}^L\dot{\bm{\omega}}(t)}\rfloor {}^I\vect{p}_L 
    + \lfloor{{}^L{\bm{\omega}(t)}}\rfloor^{2}\  {}^I\vect{p}_L
    )\\
    {}^I\bm{\omega}(t) &= {}^I_L\mat{R} {}^L\bm{\omega}(t)
    \text{,}
\end{align}
where $\lfloor \cdot \rfloor$ represents the skew-symmetric matrix of a 3D vector. Note that angular accelerations ${}^L\dot{\bm{\omega}}(t)$ are required to be computed, which makes the system complicated.
Therefore, we choose to parameterize the trajectory in IMU frame instead of LiDAR frame.

Given the raw angular velocity measurements from the IMU sensor, a series of rotational control points $\bx_q$ of the rotational trajectory can be initialized by solving the following least-square problem:
\begin{align}
\label{eq:solveq_knot}
    \hat{\bx}_q  \!&=\! \underset{\bx_q}{\operatorname{argmin}} \; r_{\omega} \notag
    \\ \!&=\! \underset{\bx_q}{\operatorname{argmin}} \;  \sum_{k} \frac{1}{2}\|{}^{\omega_k}\bomega_m \!-\!  h_{I_\omega}(t_k, \bx_q, \bx_I, \bx_{Is}) \|^2_{\Sigma_\omega} \text{,}
\end{align}
where $r_{\omega}$ is defined in \eqref{eq:gyro_loss} and here we ignore the intrinsics of IMU and the angular velocity bias of IMU.
It is important to note that ${}^{G}_{I}\bar{q}(t_0)$ is fixed to the identity quaternion during the optimization. In \eqref{eq:solveq_knot}, we try to initialize the rotational trajectory by the raw IMU measurements rather than the integrated IMU poses, since the latter are inaccurate and always affected by drifting IMU biases and noises. 

For the initialization of translational control points $\bx_p$, we first utilize the LiDAR odometry to get descrete-time LiDAR poses. Then we formulate and solve the following linear problem:
\begin{align}
    \hat{\bx}_p \!=\! \underset{\bx_p}{\operatorname{argmin}}   \sum_k
    \| {}^G_M\mathbf{T} {}^{M}\hat{\bm{p}}_{L_k} \!-\!
    {}^G_I\hat{\mathbf{R}}(t_k,\hat{\bx}_q ) {}^{I}\bm{p}_{L} \!-\! {}^G\bm{p}_I(t_k, \bx_p) \|
    \text{,}
\end{align}
where ${}^{M}\hat{\bm{p}}_{L_k}$ is the estimated position of scan at timestamp $\tau_k$ from the LiDAR odometry; ${}^G_I\hat{\mathbf{R}}(t_k, \hat{\bx}_q )$ controlled by $\hat{\bx}_q $ has been initialized in \eqref{eq:solveq_knot} and ${}^G\bm{p}_I(t_k, \bx_p)$ controlled by $\bx_p$ is the translational trajectory to be initialized.
Considering map frame and global frame are determined as the first measurements from LiDAR and IMU, respectively, ${}^G_M\mathbf{T}$ is equivalent to ${}^{I_0}_{L_0}\mathbf{T}$, that is ${}^{I}_{L}\mathbf{T}$. The initial values of ${}^{I}_{L}\mathbf{T}$ computed in Sec.~\ref{sec:extrinsic ini} are used here.
This initialization method of control points is relatively coarse, but it can effectively improve the convergence speed of the optimization problem, compared to leave $\bx_p$ uninitialized as zeros.

\subsubsection{Extrinsic Initialization}\label{sec:extrinsic ini}

Inspired by~\cite{yang2016monocular}, we initialize the extrinsic rotation by aligning two rotation sequences from LiDAR and IMU. With discrete LiDAR poses from LiDAR odometry, it is easy to get the relative rotation between two consecutive LiDAR scans, ${}^{L_{k}}_{L_{k+1}}\bar{q}$. Besides,
the relative rotation between time interval $[t_k, t_{k+1}]$ in the IMU frame can also be obtained from the initialized rotational trajectory as ${}_{I_{k+1}}^{I_k}\bar{q}={}_I^{G}\bar{q}^{-1}(t_k) \otimes{}_I^{G}\bar{q}(t_{k+1})$. The relative rotations at any $k$ from two sensors should satisfy the following equation:
\begin{align}
\label{Eq:RXXB}
    {}^{I_k}_{I_{k+1}}\bar{q}\otimes {}_L^I\bar{q} 
    = {}_L^I\bar{q}\otimes {}^{L_{k}}_{L_{k+1}}\bar{q}
    \text{.}
\end{align}
The above equation can be transferred into another equivalent representation~\cite{trawny2005indirect}:
\begin{align}
	& \mathcal{L}({}^{I_k}_{I_{k+1}}\bar{q})  {}^{I}_{L}\bar{q} =  \mathcal{R}({}^{L_k}_{L_{k+1}}\bar{q})  {}^{I}_{L}\bar{q} \notag \\
	\label{eq:initializeR}
	\Rightarrow  & \left( \mathcal{L}({}^{I_k}_{I_{k+1}}\bar{q}) -  \mathcal{R}({}^{L_k}_{L_{k+1}}\bar{q}) \right) {}^{I}_{L}\bar{q} = \mathbf{0}  \text{.}
\end{align}
For a unit quaternion $\bar{q} = \begin{bmatrix} \bq_v & q_w\end{bmatrix}^\top$ with imaginary part $\bq_v$ and real part $q_w$, $\mathcal{L}(\bar{q})$ and $\mathcal{R}(\bar{q})$ are multiplication matrices defined as follow: 
\begin{align}
		\mathcal{L}({\bar{q}})&=\begin{bmatrix}
			q_{w} \mathbf{I}_{3}-\lfloor\mathbf{q}_v\rfloor & \mathbf{q}_v \\
			-\mathbf{q}^{T}_v & q_{w}
		\end{bmatrix} \notag \\
		\mathcal{R}({\bar{q}})&=\begin{bmatrix}
			q_{w} \mathbf{I}_{3}+\lfloor\mathbf{q}_v \rfloor & \mathbf{q}_v \\
			-\mathbf{q}^{T}_v & q_{w}\end{bmatrix} \notag
			\text{.}
\end{align}
Equation~\eqref{eq:initializeR} is in a form of ${}^k_{k+1}\bA  {}^{I}_{L}\bar{q} = \mathbf{0}$. 
By stacking a series of relative rotation measurements ${}^k_{k+1}\bA$ row by row into a big matrix $\mathbf{A}$, we have $\mathbf{A} {}^{I}_{L}\bar{q} = \mathbf{0}$, then ${}^{I}_{L}\hat{\bar{q}}$ can be computed as the right unit singular vector corresponding to the smallest singular value of $\mathbf{A}$.

For the initialization of extrinsic translation ${}^I\bm{p}_L$, the initial values of ${}^I\bm{p}_L$ are simply set to zeros, and the proposed method is able to converge to reasonable values in practical experiments when enough motion excitation exists. 

\begin{figure}[t]
	\begin{subfigure}{1.0\columnwidth} 
	    \centering
		\includegraphics[width=0.8\columnwidth]{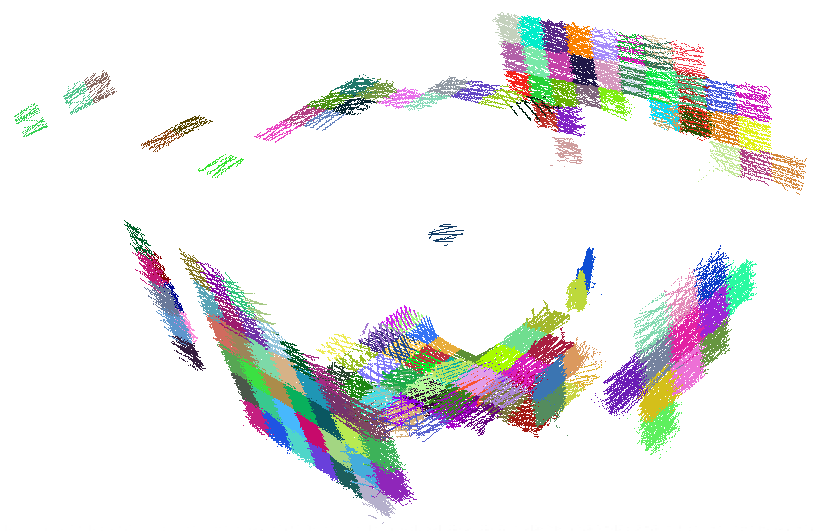}
		\caption{The surfels map in the first iteration.} 
		\label{fig:surfel_map1}
	\end{subfigure}
	\\
      \vspace{1em} 
	\begin{subfigure}{1.0\columnwidth} 
	    \centering
		\includegraphics[width=0.8\columnwidth]{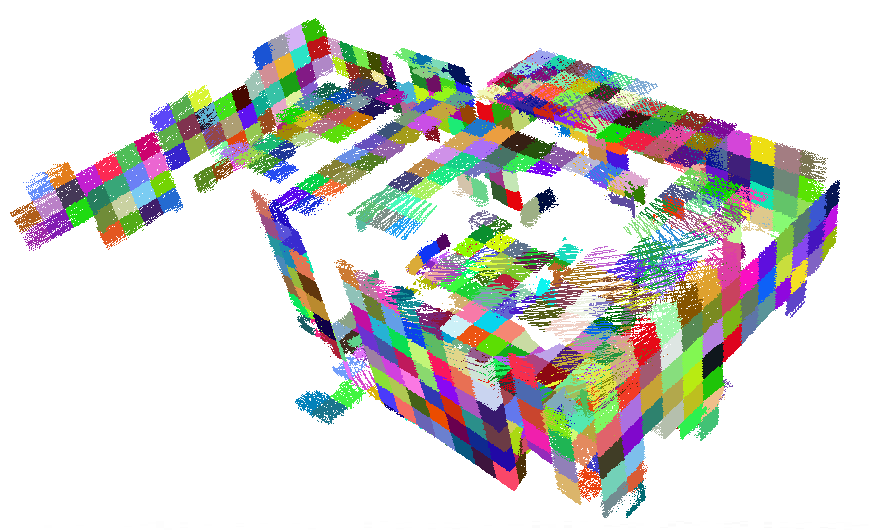}
		\caption{The surfels map in the second iteration.} 
		\label{fig:surfel_map2}
	\end{subfigure}
	\caption{Typical surfels map in an indoor scenario. The initial surfels map suffers from motion distortion with the initialized LiDAR poses. With the estimated states after just one iteration of batch optimization, the quality of refined surfels map becomes significantly better. Points in different surfels are colorized differently.} 
	\label{fig:surfel_map}
\end{figure}

\subsection{Data Association of LiDAR Measurements}
\label{sec:data_association}
With the estimated rotational extrinsics  ${}^{I}_{L}\hat{\bar{q}}$, we can transfer the rotational continuous-time curve in IMU frame into LiDAR frame and remove the rotational motion distortion in LiDAR scans. Subsequently, we perform the NDT-based LiDAR odometry again with the initial rotational guess retrieved from the rotational continuous-time curve transformed. Since LiDAR scans are partially undistorted and better initial guesses are provided, NDT-based LiDAR odometry can generate more accurate estimated poses.
With the odometry results, a LiDAR point cloud map can be constructed by assembling multiple LiDAR scans into an identical frame. We further uniformly discretize the LiDAR point cloud map into 3D cells and compute the point distribution in every 3D cell to identify surfels based on a plane-likeness coefficient~\cite{bosse2009continuous}:
\begin{equation}
	\mathcal{P}=2 \frac{\lambda_{1}-\lambda_{0}}{\lambda_{0}+\lambda_{1}+\lambda_{2}}
	\text{,}
\end{equation}
In the above equation, $\lambda_{0} \leq \lambda_{1} \leq \lambda_{2}$ are the point distribution eigen values in a 3D cell. For the cell with planar point distribution, $\mathcal{P}$ will be close to 1. The cell with plane-likeness coefficient $\mathcal{P}$ larger than a threshold is determined as a surfel cell, and the points inside this cell are used to fitting a plane. The estimated plane is parameterized by the closest point as mentioned in Sec. \ref{sec:lidar model}.
Some LiDAR points are associated with surfels if the point-to-plane distance is less than a threshold, whereas the others are discarded. 
As mentioned in Sec.~\ref{sec:lidar model}, the plane parameters are assumed to be perfect and keep constant when we perform least-squares optimization, so are the surfels map and the data associations. This is in a fashion of Expectation–maximization algorithm.
After the optimization, the surfels map, the data associations, and the plane parameters will be updated with updated estimated states, which will be detailed in Sec.~\ref{sec:refine}. 
Additionally, two practical tricks are engaged to repress the effect of outlier data associations on optimization: firstly, only the point and plane correspondence with a distance below 5 centimeters will be taken into the continuous-time batch optimization; secondly, we rely on the M-estimation~\cite{de2021review} idea and apply the Huber robust kernel to deal with outliers.

\subsection{Refinement}
\label{sec:refine}
After the continuous-time batch optimization, the estimated states including the extrinsics become more accurate. Thus, we leverage the current best estimates to remove the motion distortion in the raw LiDAR scans, rebuild the LiDAR surfels map, and update the point-to-surfel data associations. Note that results from NDT-based LiDAR odometry are only utilized at the very beginning
for initializing the LiDAR poses and LiDAR map. 
The refinement step with iterative batch-optimization will generate more accurate estimations and reliable data association.
The typical LiDAR surfels maps in the first and second iterations of the batch optimization are shown in Fig.~\ref{fig:surfel_map}. The quality of the map can be improved in a significant margin after one iteration in the batch optimization. 

As for the intrinsic calibration, a 16-beam LiDAR contains 96 parameters (6 parameters per beam), and some parameters of the first laser are fixed as a reference during the calibration process. Only the distance offset and distance scale are estimated for the first laser. Additionally, there are 15 parameters of IMU intrinsic parameters. Considering the large number of intrinsic parameters to be estimated, to prevent the system from falling into local optimum and reduce the complexity of the problem, we optimize the spatial-temporal parameters only and fix the intrinsics at the initial values firstly, and allow optimizing the intrinsic parameters a few iterations later when the estimates of the trajectory and extrinsic parameters converge.
Furthermore, we correct the raw measurements of IMU and LiDAR based on the estimated intrinsic parameters after they have gradually converged, instead of using the poor intrinsic estimates to correct the measurements at the beginning, which may hurt the calibration results. Throughout the experiments, we start to calibrate the intrinsic parameters after 2 iterations of batch optimization and correct raw measurements based on the estimated intrinsic parameters after 11 iterations.
\section{Observability-Aware LI Calibration}
\label{sec:observability}

In this section, we describe in detail the observability-aware strategies used in the proposed LI calibration 
in order to improve its robustness to degenerate motions and thus make it more easy-to-use for practitioners.

\subsection{Information-Theoretic Data Selection}\label{sec:data segment selection}

Note that some data segments collected without sufficient motion excitation or scene constraints may degrade the calibration~\cite{hausman2017observability},
and blindly processing all the available data  may result in unnecessarily high computational cost.
As such, the proposed method seeks to automatically select the most informative data segments for calibration.
In particular, to facilitate the ensuing derivations, assuming the white Gaussian noise and independent measurements, 
the NLS problem \eqref{eq:least-squares} is equivalent to the maximum likelihood estimation (MLE) problem and can be re-formulated as follows (see~\cite{kaess2012isam2,eckenhoff2019closed}):
\begin{align}
	\label{eq:leastsaures}
	\hat{\bx} = \underset{\bx}{\operatorname{argmin}} \sum_{i} \frac{1}{2} \|\be_i(\bx)\|^2_{\Sigma_i}
\end{align}
where $\be_i$ denotes the residual induced from the $i$-th measurement with covariance ${\Sigma_i}$. 
An iterative  method such as Gaussian-Newton and Levenberg-Marquard is often used to solve this problem.
That is, at each iteration, by linearizing the above nonlinear system at the current estimate
$\hat{\bx}$, $\be_i(\bx) = \be_i(\hat{\bx}) + \bJ_i \delta \bx$, with the Jacobian matrix $\bJ_i$,
we have the following linear least-squares problem with respect to the update increment (or error state estimate) $\delta \hat{\bx}$:
\begin{align}
 	\delta\hat{\bx} = \underset{\delta\bx}{\operatorname{argmin}} \sum_{i} \frac{1}{2} \|\be_i(\hat{\bx}) + \bJ_i \delta \bx \|^2_{\Sigma_i}
\end{align}
The optimal solution is given by the following normal equation:
\begin{align}
	\underbrace{ \left( \sum_{i} \bJ^\top_i {\Sigma^{-1}_i} \bJ_i  \right) }_{\mathbf A} \delta \bx &= \underbrace{ - \sum_i \bJ_i^{\top} {\Sigma^{-1}_i} \be_i(\hat{\bx}) }_{\mathbf b}
	\label{eq:Ax=b}
\end{align}
Once $\delta\hat{\bx}$ is computed, we update the estimate as $\hat{\bx} \leftarrow \hat{\bx} \boxplus \delta\bx$,
where $\boxplus$ denotes the addition operation on manifold. 

Note that $\bA_{k\times k}$ in \eqref{eq:Ax=b} is the Fisher information matrix of the MLE  (or NLS) problem~\eqref{eq:leastsaures}
and encapsulates all the information contained in the measurements.
It is also important to note that 
$\bA$ might be rank deficient, for example, due to the lack of motion excitation and/or environmental constraints (see \cite{Zuo2020IROS}). 
It thus is critical to collect informative data ensuring  full-rank $\bA$ (or observability~\cite{Jauffret2007TAES}).
To this end, we first partition the entire collected data into multiple segments with constant-time length. 
We then perform Singular Value Decomposition (SVD) of $\bA$ for each segment:
\begin{align}
	\label{eq:svd}
	\bA = \bU \bS \bU^\top
\end{align}
where $\bS = \operatorname{diag}(\sigma_1, \sigma_2, \cdots, \sigma_k)$ is the diagonal matrix of singular values in decreasing order, and $\bU = \begin{bmatrix} \bu_1, \bu_2, \cdots, \bu_k \end{bmatrix}$ 
is an orthogonal matrix. As in \cite{Jauffret2007TAES}, we employ the information metric (or observability index) of an individual segment based on the minimum singular value -- that is, a larger one implies a more informative segment. With that, for a specific sequence of collected data, we first evaluate the information metric of each segment and then select the most informative ones for LI calibration.

\subsection{Observability-Aware Update}
\label{sec:tsvd}

While we have tried to collect the most informative data for calibration,
it still might be the case that the selected data cannot guarantee observability and thus some states cannot be estimated.
For example, when a ground vehicle moves on planar surfaces  in an urban environment, 
the lack of pitch and roll rotation and translation on the vertical direction often leads to part of states unobservable.
As shown in~\cite{Zuo2020IROS},  one-axis rotation is a so-called degenerate motion  causing the translation of extrinsics ${}^I\bp_L$ partially unobservable.
If this unobservability issue is not properly addressed, the calibration performance would be largely degraded.

To make the proposed LI calibration robust to (weak) unobservability,
we explicitly enforce  observability constraints during each iteration;
that is, only updating the states lying on the observable directions.
To this end, we employ 
the truncated SVD (TSVD)~\cite{hansen1987truncatedsvd,maye2016online} to perform low-rank approximation of the information matrix, 
i.e., selecting the singular values $(\sigma_1, \sigma_2, \cdots,\sigma_l)$ over an information threshold $\epsilon$ that is a design choice.
With this observability-assurance information matrix and SVD-based matrix pseudo-inverse, we have the following solution [see \eqref{eq:Ax=b}]:
\begin{align}
	\delta \hat{\bx} = \sum^{l}_{i=1} \frac{\bu^\top_i \mathbf{b} \bu_i}{\sigma_i}.
	\label{eq:tsvd}
\end{align}
As a result, as compared to solving for $\delta \hat{\bx}$ by using all the singular values of the full information matrix -- which may erroneously introduce observable directions due to numerical issues --
we here explicitly avoid updating the unobservable (or weakly observable)  parameters, thus conservatively ensuring observability. 

\section{Simulation Results}
\label{sec:simu_experiment}

To validate the performance of the proposed continuous-time batch optimization based LI calibration method, 
we conduct a series of simulations with ground truth of all estimated parameters to examine the accuracy of intrinsic and extrinsic calibration results,
and also numerically study the sensitivity of the system to the time offset in simulations.

\begin{figure}[t]
	\centering
	\includegraphics[width=1.0\columnwidth]{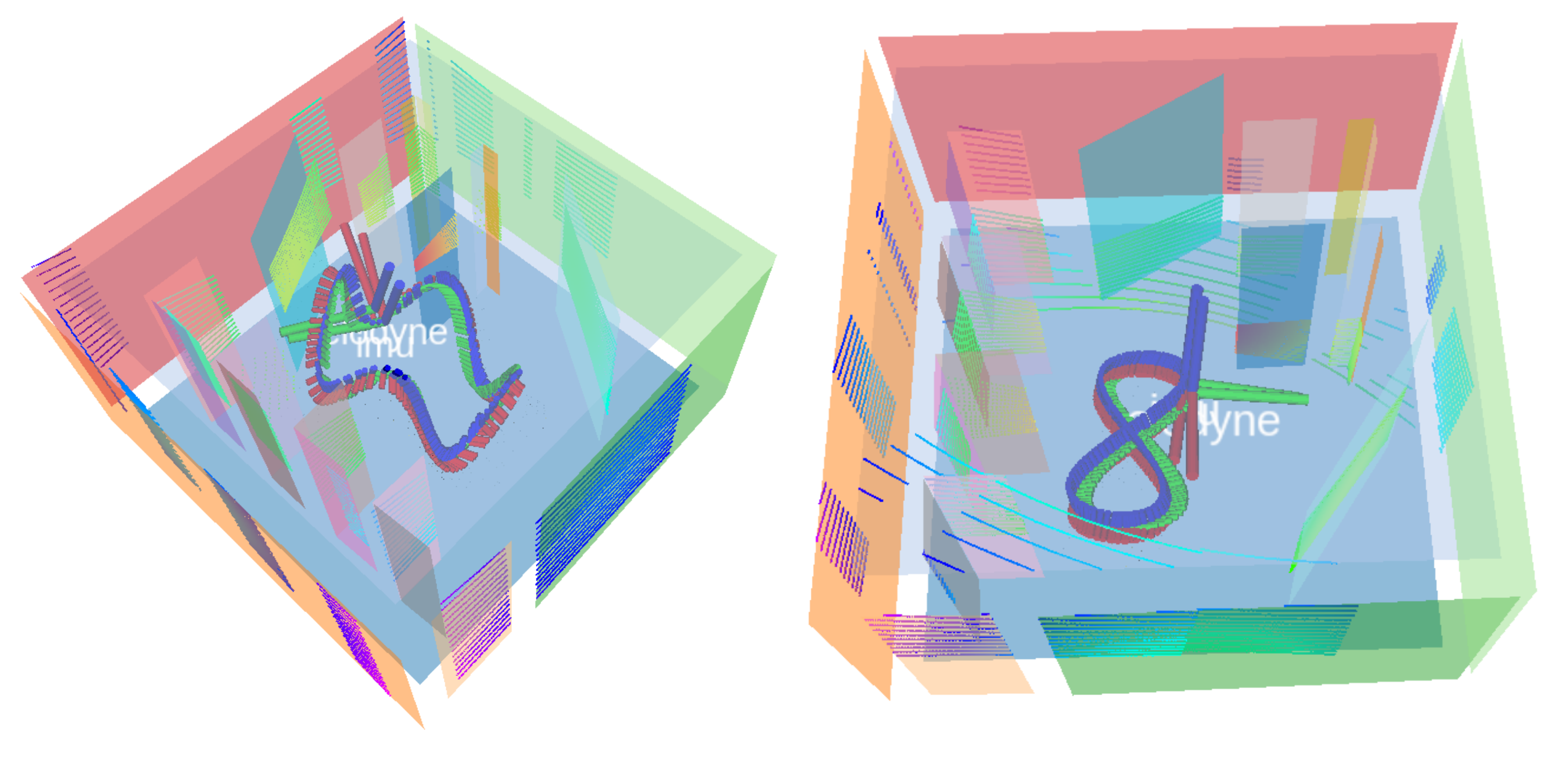}
	\caption{The simulated environmental setup filled with planes in different colors. The colored points indicate a simulated 3D LiDAR scan. Left: The sinusoidal trajectory for testing calibration in full excitation motion. Right: Figure-8-shape trajectory for testing calibration in degenerate motion.}
	\label{fig:simu_env}
\end{figure}

\begin{table}[t]
\centering
\caption{Standard deviations for generating random LiDAR and IMU intrinsics in simulation experiments.}
\begin{adjustbox}{width=\columnwidth,center}
	\begin{tabular}{cccc} 
    \toprule
    \textbf{Parameter}                                   & \textbf{Value} & \textbf{Parameter}                    & \textbf{Value} \\ \hline
    Elevation angle correction $\delta\phi$ (${}^\circ$) & 0.1            & Vertical spatial offset $V$ ($m$)     & 0.005          \\
    Azimuth angle correction $\delta\theta$ (${}^\circ$) & 0.05           & Horizontal spatial offset $H$ ($m$)   & 0.005          \\
    Range scale factor$s$ (\%)                           & 0.005          & Range offset $\delta \rho$ ($m$)      & 0.01           \\
    Gyro. Non-linearity (\%)                             & 0.01           & Accel. Non-linearity (\%)             & 0.1            \\
    Gyro. Non-orthogonality (${}^\circ$)                 & 0.05           & Accel. Non-orthogonality (${}^\circ$) & 0.05           \\ 
	\bottomrule
	\end{tabular}
\end{adjustbox}
\label{tab:sensor_intrinsic_setting}
\end{table}

For the results presented below, 
we simulate an Xsens-100 IMU\footnote{\url{https://www.xsens.com/products/mti-100-series/}} and a 16-beam 3D LiDAR VLP-16\footnote{\url{https://velodyneLiDAR.com/vlp-16.html}}. The synthetic noises in measurement are consistent with the real-world sensors. IMU measurements are reported at 400Hz. LiDAR scans are received at  10Hz and the laser rays cover 360${}^\circ$ horizontally and $\pm$15${}^\circ$ vertically. The motion distortion effect is also simulated to match reality.
The intrinsic parameters of IMU and LiDAR are randomly generated from normal distributions with standard deviations provided in Table~\ref{tab:sensor_intrinsic_setting}. 
The extrinsic parameters from LiDAR to IMU are set as $[0.3,0.15,0.05]^{\top}$ meter in translation and $[1,2,5]^{\top}$ degree in rotation in all simulation experiments. Note that the extrinsic translation has a minimum component of 0.05 meters and a maximum component of 0.3 meters to verify the calibration accuracy of the proposed method for different magnitudes of the translation.

For the simulated trajectories, a sinusoidal trajectory and a figure-8-shape trajectory are introduced. The sinusoidal trajectory is formulated as: 
\begin{equation}
\left\{
\begin{array}{l}
     p_x(t) = 2.0 \cdot \cos(\frac{\pi}{5}\cdot t) + 5 \\
     p_y(t) = 1.5 \cdot \sin(\frac{\pi}{5}\cdot t) + 5\\
     p_z(t) = 0.8 \cdot \cos(\frac{4\pi}{5}\cdot t)  + 5\\
     r_x(t) = 0.4 \cdot \cos(t)  \\
     r_y(t) = 0.6 \cdot \sin(t) \\
     r_z(t) = 0.7\cdot t 
\end{array}
\right.
\label{eq:simu_traj1}
\end{equation}
and the trajectory of figure-8-shape is as follows: 
\begin{equation}
\left\{
\begin{array}{l}
     p_x(t) = 2.0 \cdot \cos(\frac{\pi}{5}\cdot t)  \\
     p_y(t) = 1.5 \cdot \sin(\frac{\pi}{5}\cdot t)\cdot\cos(\frac{\pi}{5}\cdot t) + 5\\
     p_z(t) = 2  \\
     r_x(t) = r_y(t) = 0 \\
     r_z(t) = 0.4 \cdot \sin(t)  \\
\end{array}
\right.
\label{eq:simu_traj2}
\end{equation}
where $t\in[0,10]$, $\bp(t)=[p_x(t)\ \  p_y(t)\ \ p_z(t)]^{\top}$ is the translational trajectory in meter and $\vect{r}_{\text{euler}}(t) = [r_z(t)\ \ r_y(t)\ \ r_z(t)]^{\top}$ represents the euler angles in rad. 
The sinusoidal trajectory is with fully-excited motions, while the figure-8-shape trajectory is a simulation of planar motion, where the robot moves on the $x$-$y$ plane, and orientations change along $z$-axis only. 
For both simulated trajectories, it is straightforward to derive the linear acceleration and angular velocity at any time  in order to generate synthetic  IMU readings. 
Note that in each experiment, we perform 10 Monte Carlo experiments with the same time span of 10 seconds.

% Introduce environment
The simulated environment is a room constructed with many planes and a volume of $12\times10\times10 \ m^3$ as shown in Fig.~\ref{fig:simu_env}, where the two types of trajectories are also plotted.  The sinusoidal trajectory evaluates the intrinsic and extrinsic calibration in full motion excitation while the figure-8-shape trajectory for analyzing the extrinsic calibration performance in degenerate motion.

To evaluate the impact of LiDAR intrinsic calibration on the accuracy of map reconstruction, we introduce the \textit{mean map entropy} (MME) metric~\cite{droeschel2014local}, which evaluates the sharpness of a map. The entropy $h$ for a map point $\mathbf{p}_k$ is calculated by 
\begin{align}
    h(\mathbf{p}_k) = \frac{1}{2} \ln{|2\pi e \Sigma(\mathbf{p_k})|}
\end{align}
where $\Sigma(\mathbf{p_k})$ is the sample covariance of map points within a local radius $r$ around $\mathbf{p_k}$. Local radius $r$ is 0.3 meters in our experimental evaluations. The MME $H(Q)$ is averaged over all map points $Q$:
\begin{align}
    H(Q) = \frac{1}{n} \sum_{k=1}^{n}h(\mathbf{p_k})
    \text{.}
\end{align}
The smaller the MME value  the sharper  the map is (i.e., the better map reconstruction). 

\begin{table}[t]
\centering
\caption{Results of 10 Monte-Carlo simulations with the sinusoidal trajectory: mean and standard deviation of the extrinsic calibration results with and without calibrating the intrinsics of LiDAR and IMU. Note that the best results are marked in bold.}
%\resizebox{\columnwidth}{!}{%
    \begin{tabular}{rcr@{}c@{}lr@{}c@{}l}
    \toprule
    Parameter   & Gt & \multicolumn{3}{c}{W/  Calib. Intr.} & \multicolumn{3}{c}{W/O Calib. Intr.} \\
    \midrule
    $p_x$ (cm)     & 30 & \textbf{29.81}   & ~$\pm$~  & 0.54  & 35.15   & ~$\pm$~  & 0.51  \\
    $p_y$ (cm)     & 15 & \textbf{14.86}   & ~$\pm$~  & 0.25  & 22.14   & ~$\pm$~  & 0.73  \\
    $p_z$ (cm)     & 5  & \textbf{4.68}    & ~$\pm$~  & 0.28  & 1.67    & ~$\pm$~  & 0.14  \\
    roll (deg)  & 1  & \textbf{1.16}    & ~$\pm$~  & 0.02  & 1.97    & ~$\pm$~  & 0.02  \\
    pitch (deg) & 2  & \textbf{1.91}    & ~$\pm$~  & 0.02  & 2.41    & ~$\pm$~  & 0.03  \\
    yaw (deg)   & 5  & \textbf{5.02}    & ~$\pm$~  & 0.05  & 5.44    & ~$\pm$~  & 0.05  \\
    \bottomrule
    \end{tabular}
%}
\label{tab:simu_extrinsic}
\end{table}

\begin{table}[t]
\centering
\caption{Results of simulations with the sinusoidal trajectory: time offset calibrations.}
\resizebox{\columnwidth}{!}{%
\begin{tabular}{cccccccc}
\toprule
              & \multicolumn{7}{c}{Time offset (ms)}                 \\
\midrule
\rowcolor[HTML]{EFEFEF} 
    True value& 1    & 2     & 3     & 5     & 8     & 12    & 21    \\ %\hline
    Estimated & 1.01 & 1.86  & 2.77  & 4.79  & 7.87  & 12.94 & 20.63 \\
    \rowcolor[HTML]{EFEFEF} %\hline
    Error     & 0.01 & -0.14 & -0.23 & -0.21 & -0.13 & -0.06 & -0.37 \\
\bottomrule
\end{tabular}
}
\label{tab:simu_t_offset}
\end{table}

\begin{figure*}[th]
	\centering
	\includegraphics[width=1.0\textwidth]{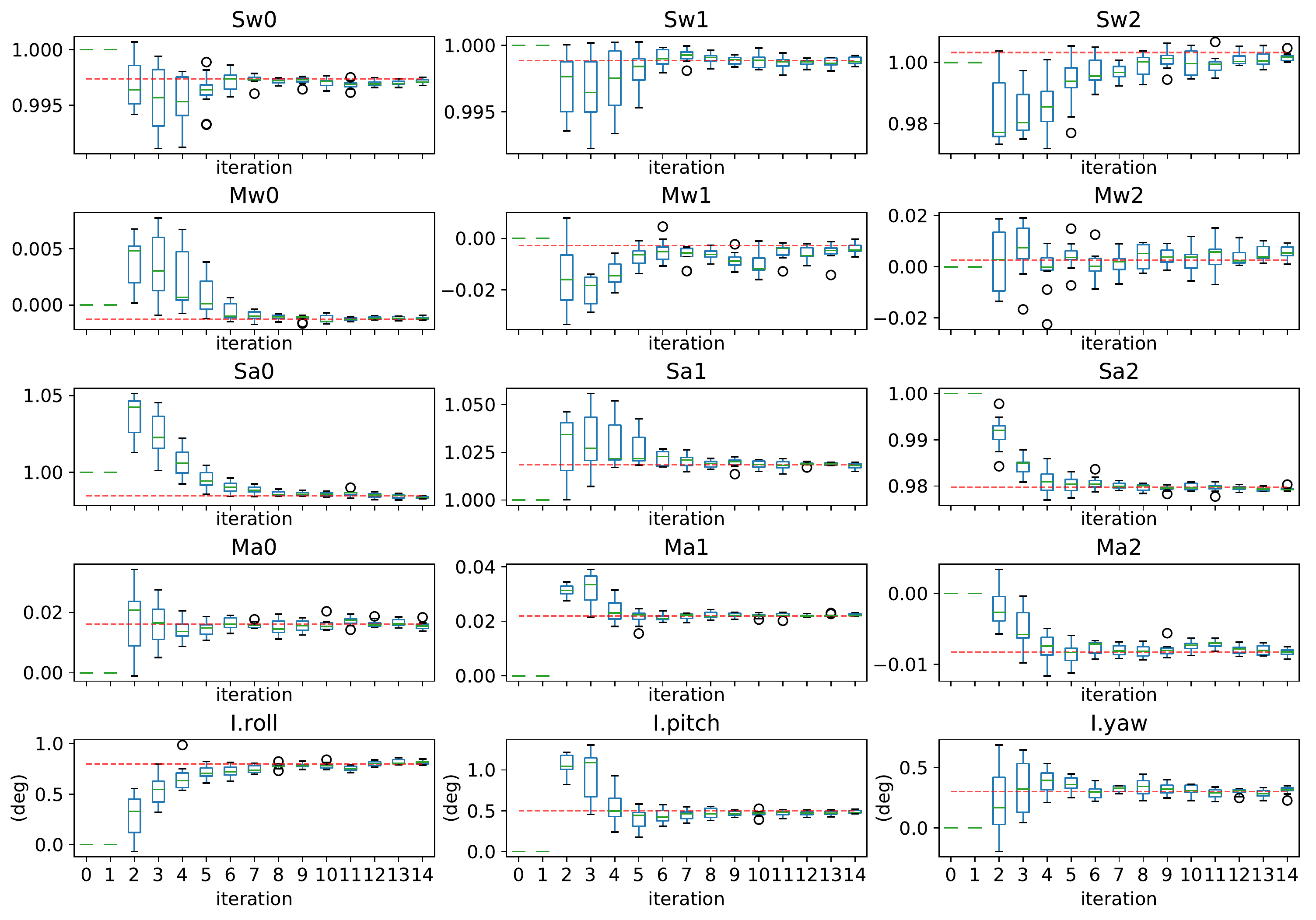}
	\caption{Results of 10 Monte-Carlo simulations with the sinusoidal trajectory: the IMU intrinsic parameters change with the increased number of iterations. The ground truth is denoted with the red dashed line. Each row from top to bottom represents scale and misalignment parameters of gyroscope, and scale and misalignment parameters of accelerometer, and euler angles of ${}^\omega_I\bR$ in \eqref{eq:imu_w_model}, respectively.
	}
	\label{fig:simu_IMU_intrinsic}
\end{figure*}

\begin{figure*}[th]
	\centering
	\includegraphics[width=1.0\textwidth]{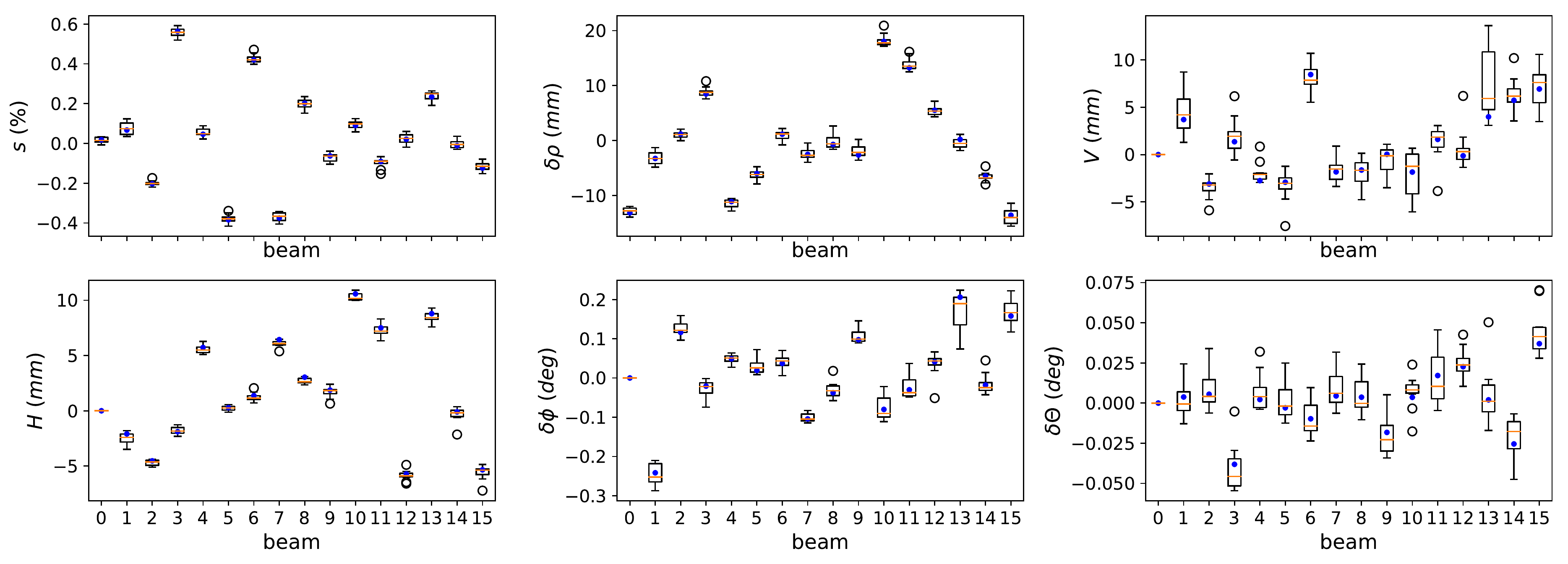}
	\caption{Results of 10 Monte-Carlo simulations with the sinusoidal trajectory: the intrinsic calibration results of a simulated 16-beam 3D LiDAR. For the first laser, only the distance offset and distance scale are estimated while the other parameters are fixed as a reference during the calibration process. The blue dots are the ground truth and red lines are the mean of each parameter.}
	\label{fig:simu_lidar_intrinsic}
\end{figure*}

\subsection{Calibration With Full Motion Excitation}

\subsubsection{{Spatial-Temporal Calibration}}

The extrinsic calibration results with and without calibrating the intrinsic parameters of LiDAR and IMU in 10 simulated sequences are shown in Table~\ref{tab:simu_extrinsic}.  We can find that with the awareness of intrinsic parameters, the proposed method OA-LICalib achieves an accuracy of 0.40 cm in translation and 0.18 degree in rotation, and the calibration errors extremely increase without  intrinsic calibrations of LiDAR and IMU. %
It also confirms OA-LICalib's capability of calibrating extrinsic parameters with components at different magnitudes, e.g., translation varies from 5cm to 30 cm, and rotation varies from 1 degree to 5 degrees in this experiment. It is also interesting to note that the magnitudes of the rotational and translational trajectories in \eqref{eq:simu_traj1} affect the accuracy of the extrinsic calibration results.
In the $x$-axis translation and $z$-axis rotation, OA-LICalib gives calibrated parameters with much lower relative error.
This phenomenon agrees with the observability analysis in~\cite{kelly2011visual} that calibration trajectory with strong motion excitation, i.e., large accelerations and angular velocities, can lead to more accurate calibrations of extrinsic parameters than weal excitation.

Furthermore, the temporal calibration results in 7 trails are shown in Table~\ref{tab:simu_t_offset}. Note that time offsets vary from 1 ms to 21 ms, are manually added to one of the simulated sequences and calibrated. For the precision of temporal calibration, the proposed method can achieve millisecond-level accuracy. 

\begin{table}[t]
\centering
\caption{Results of 10 Monte-Carlo simulations with the sinusoidal trajectory:  mean and error of IMU intrinsic parameters after 14 iterations.}
\begin{tabular}{@{}ccr@{}c@{}lc@{}}
\toprule
Parameter & Gt & \multicolumn{3}{c}{Estimated} & Error \\ \midrule
Sw0 & 0.99738 & 0.99714 &  {$\pm$} &  {0.00020} & -0.00024 \\
Sw1 & 0.99885 & 0.99882 &  {$\pm$} &  {0.00031} & -0.00003 \\
Sw2 & 1.00329 & 1.00169 &  {$\pm$} &  {0.00132} & -0.00160 \\
Sa0 & 0.98487 & 0.98374 &  {$\pm$} &  {0.00071} & -0.00113 \\
Sa1 & 1.01851 & 1.01780 &  {$\pm$} &  {0.00164} & -0.00071 \\
Sa2 & 0.97975 & 0.97936 &  {$\pm$} &  {0.00037} & -0.00039 \\ \midrule
Mw0 & -0.00127 & -0.00117 &  {$\pm$} &  {0.00015} & 0.00010 \\
Mw1 & -0.00279 & -0.00388 &  {$\pm$} &  {0.00193} & -0.00109 \\
Mw2 & 0.00254 & 0.00577 &  {$\pm$} &  {0.00254} & 0.00323 \\
Ma0 & 0.01607 & 0.01553 &  {$\pm$} &  {0.00129} & -0.00054 \\
Ma1 & 0.02197 & 0.02238 &  {$\pm$} &  {0.00049} & 0.00041 \\
Ma2 & -0.00825 & -0.00825 &  {$\pm$} &  {0.00052} & 0.00001 \\ \midrule
I.roll (deg) & 0.80000 & 0.81411 &  {$\pm$} &  {0.01924} & 0.01411 \\
I.pitch (deg) & 0.50000 & 0.48720 &  {$\pm$} &  {0.02066} & -0.01280 \\
I.yaw (deg) & 0.30000 & 0.31130 &  {$\pm$} &  {0.03401} & 0.01130 \\ \bottomrule
\end{tabular}
\label{tab:simu_imu_intrinsic}
\end{table}

\begin{table}[t]
\centering
\caption{Results of 10 Monte-Carlo simulations with the sinusoidal trajectory: the averaged absolute errors of the estimated LiDAR intrinsic parameters after 14 iterations. 
}
\begin{tabular}{@{}cr@{}c@{}llr@{}c@{}l@{}}
\toprule
Parameter & \multicolumn{3}{c}{Error} & \multicolumn{1}{c}{Parameter} & \multicolumn{3}{c}{Error} \\ \midrule
$s\ (\%)$ & 0.0058 &  {$\pm$} &  {0.0023} & $\delta \rho \ (mm)$ & 0.3048 &  {$\pm$} &  {0.2167} \\
$\delta \phi \ (deg)$ & 0.0075 &  {$\pm$} &  {0.0081} & $V\ (mm)$ & 0.6383 &  {$\pm$} &  {0.7838} \\
$\delta \theta \ (deg)$ & 0.0029 &  {$\pm$} &  {0.0019} & $H\ (mm)$ & 0.2111 &  {$\pm$} &  {0.1106} \\ \bottomrule
\end{tabular}
\label{tab:simu_lidar_intrinsic}
\end{table}

\begin{figure}[ht]
	\centering
	\includegraphics[width=\columnwidth]{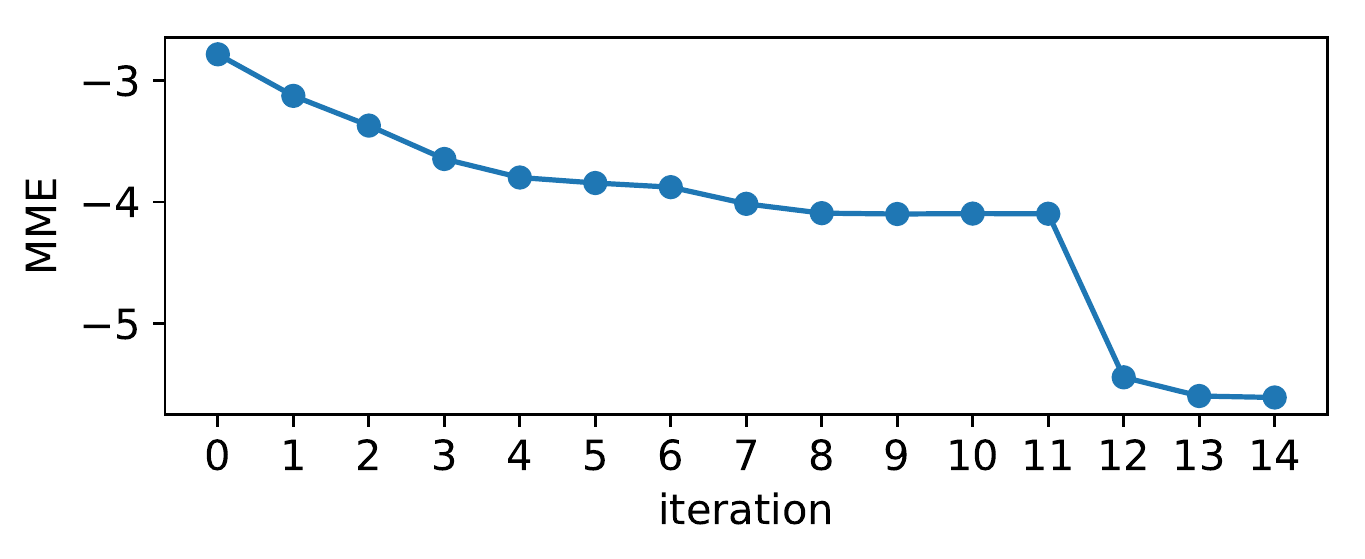}
	\caption{The MME evaluation of the LiDAR map quality along with the number of iterations in a typical simulated experiment. The evaluated point cloud map is initially assembled with the poses from NDT-based LiDAR odometry, then assembled with iteratively refined poses and intrinsics from the continuous-time batch optimization.}
	\label{fig:simu_map_MME}
\end{figure}

\begin{figure}[t]
	\centering
	\includegraphics[width=\columnwidth]{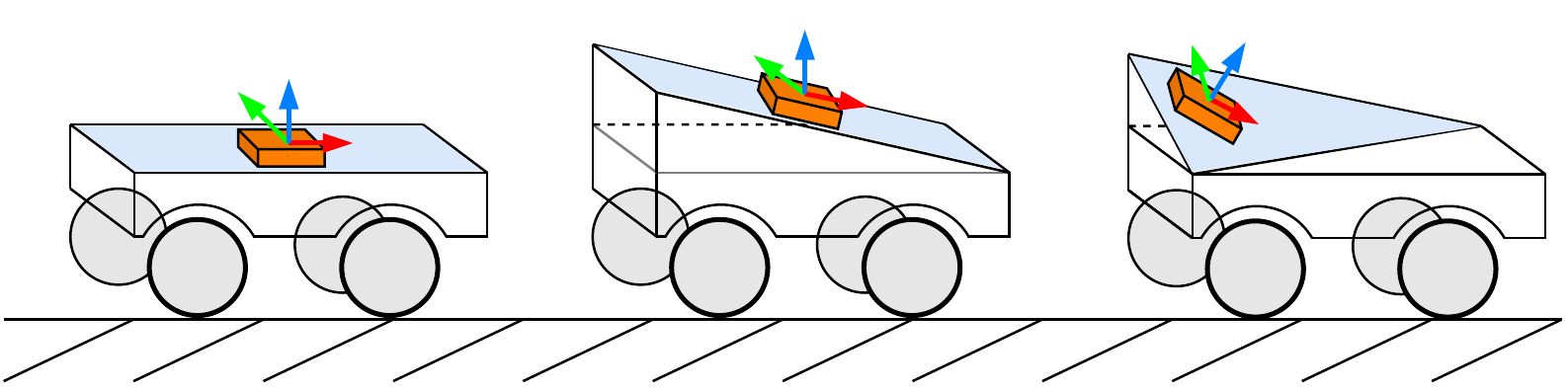}
	\caption{Simulation to examine observability awareness: three mounting setups with different relative orientations between the robot and the LiDAR-IMU sensor rig are tested (LiDAR and IMU are rigidly connected, while LiDAR is not shown in this figure). From left to right: case A, case B, case C. Among these three cases, the robots move in a shared trajectory of figure-8 shape as \eqref{eq:simu_traj2}, while the unobservable extrinsic directions in local IMU frames are different. 
    }
	\label{fig:simu_degenerate_dir}
\end{figure}

\begin{table}[t]
    \centering
    \caption{Results of simulation with the figure-8 trajectory: the ground truth and the automatically-detected unobservable extrinsic directions under three mounting cases are shown in Fig.~\ref{fig:simu_degenerate_dir}. The head part corresponds to extrinsic rotation and the tail part corresponds to extrinsic translation.}
    \resizebox{\columnwidth}{!}{%
    \begin{tabular}{@{}cccccccc@{}}
    \toprule
    Case                    &                                  & \multicolumn{6}{c}{Degenerate direction}                                                                                                                                                                     \\ \midrule
                            & Truth                          & 0.00000                          & 0.00000                         & 0.00000                          & 0.00000                         & 0.00000                          & 1.00000                         \\
    \multirow{-2}{*}{Case A} & \cellcolor[HTML]{EFEFEF}Detected & \cellcolor[HTML]{EFEFEF}-0.00001 & \cellcolor[HTML]{EFEFEF}0.00000 & \cellcolor[HTML]{EFEFEF}0.00005  & \cellcolor[HTML]{EFEFEF}0.00165 & \cellcolor[HTML]{EFEFEF}-0.00110 & \cellcolor[HTML]{EFEFEF}1.00000 \\
                            & Truth                          & 0.00000                          & 0.00000                         & 0.00000                          & 0.50000                         & 0.00000                          & 0.86603                         \\
    \multirow{-2}{*}{Case B} & \cellcolor[HTML]{EFEFEF}Detected & \cellcolor[HTML]{EFEFEF}0.00000  & \cellcolor[HTML]{EFEFEF}0.00000 & \cellcolor[HTML]{EFEFEF}-0.00001 & \cellcolor[HTML]{EFEFEF}0.50117 & \cellcolor[HTML]{EFEFEF}-0.00023 & \cellcolor[HTML]{EFEFEF}0.86535 \\
                            & Truth                          & 0.00000                          & 0.00000                         & 0.00000                          & 0.50000                         & 0.43301                          & 0.75000                         \\
    \multirow{-2}{*}{Case C} & \cellcolor[HTML]{EFEFEF}Detected & \cellcolor[HTML]{EFEFEF}0.00006  & \cellcolor[HTML]{EFEFEF}0.00004 & \cellcolor[HTML]{EFEFEF}0.00007  & \cellcolor[HTML]{EFEFEF}0.50070 & \cellcolor[HTML]{EFEFEF}0.43296  & \cellcolor[HTML]{EFEFEF}0.74957 \\ \bottomrule
    \end{tabular}
    }
    \label{tab:simu_fig8_dir}
\end{table}

\begin{table}[t]
\centering
\caption{Results of 10 Monte-Carlo simulations with the figure-8 trajectory: mean of the estimated extrinsic parameters in three mounting cases with or without observability-aware calibration.}
\resizebox{\columnwidth}{!}{%
\begin{tabular}{|cccccccccc|}
\hline
 &  & \multicolumn{3}{c}{\cellcolor[HTML]{E0EFD4}${}^I\bp_L$ (cm)} & \multicolumn{3}{c}{\cellcolor[HTML]{E0EFD4}${}^I_L\bar{q}$ in Euler (deg)} & \multicolumn{2}{c|}{\cellcolor[HTML]{EECC89}RMSE} \\ \cline{3-10} 
\multirow{-2}{*}{Seq.} & \multirow{-2}{*}{Method} & \cellcolor[HTML]{E0EFD4}$p_x$ & \cellcolor[HTML]{E0EFD4}$p_y$ & \cellcolor[HTML]{E0EFD4}$p_z$ & \cellcolor[HTML]{E0EFD4}roll & \cellcolor[HTML]{E0EFD4}pitch & \cellcolor[HTML]{E0EFD4}yaw & \cellcolor[HTML]{EECC89}Pos (cm) & \cellcolor[HTML]{EECC89}Rot (deg) \\ \hline
\multicolumn{2}{|c}{Groundtruth} & 30.00 & 15.00 & 5.00 & 1.00 & 2.00 & 5.00 &  &  \\ \hline
 & \cellcolor[HTML]{EFEFEF}W/O Obser. & \cellcolor[HTML]{DDDDDD}27.24 & \cellcolor[HTML]{DDDDDD}11.49 & \cellcolor[HTML]{DDDDDD}7.17 & \cellcolor[HTML]{DDDDDD}0.85 & \cellcolor[HTML]{DDDDDD}2.40 & \cellcolor[HTML]{DDDDDD}4.83 & \cellcolor[HTML]{DDDDDD}2.87 & \cellcolor[HTML]{DDDDDD}0.27 \\
\multirow{-2}{*}{Case A} & W Obser. & 30.97 & 12.09 & 8.05 & 0.95 & 2.44 & 4.89 & \textbf{2.50} & \textbf{0.26} \\ \hline
 & \cellcolor[HTML]{EFEFEF}W/O Obser. & \cellcolor[HTML]{DDDDDD}26.50 & \cellcolor[HTML]{DDDDDD}11.53 & \cellcolor[HTML]{DDDDDD}6.88 & \cellcolor[HTML]{DDDDDD}0.76 & \cellcolor[HTML]{DDDDDD}2.27 & \cellcolor[HTML]{DDDDDD}4.88 & \cellcolor[HTML]{DDDDDD}3.04 & \cellcolor[HTML]{DDDDDD}0.22 \\
\multirow{-2}{*}{Case B} & W Obser. & 29.73 & 11.94 & 5.69 & 1.01 & 2.27 & 4.81 & \textbf{1.82} & \textbf{0.19} \\ \hline
 & \cellcolor[HTML]{EFEFEF}W/O Obser. & \cellcolor[HTML]{DDDDDD}26.32 & \cellcolor[HTML]{DDDDDD}11.20 & \cellcolor[HTML]{DDDDDD}6.03 & \cellcolor[HTML]{DDDDDD}0.08 & \cellcolor[HTML]{DDDDDD}2.28 & \cellcolor[HTML]{DDDDDD}4.86 & \cellcolor[HTML]{DDDDDD}3.11 & \cellcolor[HTML]{DDDDDD}0.56 \\
\multirow{-2}{*}{Case C} & W Obser. & 29.67 & 10.42 & 5.56 & 0.18 & 2.28 & 4.88 & \textbf{2.67} & \textbf{0.50} \\ \hline
\end{tabular}
}
\label{tab:simu_fig8}
\end{table}

\subsubsection{{Intrinsic Calibration}}
Figure~\ref{fig:simu_IMU_intrinsic} illustrates the convergence process of IMU intrinsic parameters along with iterations and visualizes the distribution of calibration results for 10 simulated trials. Note that I.roll, I.pitch and I.yaw are the corresponding Euler angles of ${}^\omega_I\bR$ in \eqref{eq:imu_w_model}.
The intrinsic parameters remain unchanged during the first iteration and start to be optimized in the second iteration, and the results gradually stabilize after about 8 iterations. Since the 11th iteration, the raw measurements are corrected based on the estimated intrinsic parameters (as mentioned in Sec.~\ref{sec:refine}). We can see that the estimated intrinsic parameters finally converge to near the true value, and the final estimated intrinsic results are summarized in Table~\ref{tab:simu_imu_intrinsic}. The experiment indicates that the proposed method has high accuracy and repeatability in intrinsic calibration.

The final calibration results of the simulated 16-beam LiDAR intrinsic parameters in 10 trails are shown in Fig.~\ref{fig:simu_lidar_intrinsic}. Besides, the averaged absolute errors of the estimated LiDAR intrinsic parameters in 6 types are summarized in Table~\ref{tab:simu_lidar_intrinsic}.
The proposed method can identify six types of intrinsic parameters of multi-beam LiDAR and estimate accurate results. We further evaluate the quality of LiDAR point cloud map which is affected by both the estimated trajectory and the calibrated parameters. 
Figure~\ref{fig:simu_map_MME} shows the MME metric in a typical simulated experiment. 
The evaluated point clouds consisting of around $215k$ points are assembled by the poses from NDT-based LiDAR odometry firstly, and refined poses from the continuous-time batch optimization.
The map becomes more accurate with increased iterations due to the batch optimization over the continuous-time trajectory and calibrated parameters. Notably, a significant improvement in map accuracy after the 11th iteration due to the correction of the raw LiDAR points based on the current best estimation of LiDAR intrinsics. It is obvious that intrinsic parameters calibration significantly affects both the LiDAR map quality and the extrinsic calibration. Therefore, we advocate calibrating the intrinsic parameters to enable highly accurate LIDAR localization and mapping.

\subsection{Calibration Under Degenerate Motion}
In the same simulation environment with figure-8-shape trajectory, shown in the right picture of Fig.~\ref{fig:simu_env}, we investigate the extrinsic calibration accuracy of OA-LICalib in the degenerate case. With the trajectory in \eqref{eq:simu_traj2}, where rotation happens along the $z$-axis only, the  LiDAR-IMU extrinsic translation along the rotation axis is unobservable according to the analysis in~\cite{Zuo2020IROS}.
In order to examine whether the proposed OA-LICalib is able to automatically identify the unobservable directions, we also simulate three different mounting cases (A, B, C) as shown in Fig.~\ref{fig:simu_degenerate_dir}, where the LiDAR-IMU sensor rig is rigidly attached to the robot platform at three different relative orientations.
Among the three cases, the robots move in the same trajectory (as Eq.~\eqref{eq:simu_traj2}), thus the unobservable directions of the extrinsic translation are the identical vertical axis in the robot coordinate frame.
However, due to the different relative rotations between robot platform and the LiDAR-IMU sensor rigs, the unobservable directions of the extrinsic translation, ${}^{I}\mathbf{p}_{L}$, are different when represented in local IMU frame.

Table~\ref{tab:simu_fig8_dir} summarizes the unobservable direction of the three mounting cases in Fig.~\ref{fig:simu_degenerate_dir}. For case A where the IMU $z$-axis is aligned with the vertical axis, the unobservable direction is purely in the $z$-axis of extrinsic translation, ${}^{I}\mathbf{p}_{L}$. For a typical trial, OA-LICalib identifies the singular value of the information matrix corresponding to extrinsics as $[41760,\ 22193,\ 3525,\ 491,\ 295,\ 0.000]^{\top}$, and it's obvious that there exists one unobservable direction. The eigen vectors corresponding to the minimum singular value (deemed as the detected unobservable direction) in the three cases are also summarized in Table~\ref{tab:simu_fig8_dir}. We can see that the detected unobservable directions are consistent with the truths.
To verify the effectiveness of the observability scheme in Section.~\ref{sec:tsvd}, we conduct the ablation experiments on extrinsic calibration by starting from perturbed initial extrinsic parameters (with perturbations of around 3 degrees and 3 centimeters on rotations and translations, respectively).
Table~\ref{tab:simu_fig8} shows the extrinsic calibration results ($\{ {}^I_L\bar{q}, \,  {}^I \bp_{L} \}$) of 10 Monte-Carlo trials, and the Root Mean Square Error (RMSE) is calculated by comparing the mean calibration results to the ground truths. 
OA-LICalib with observability awareness keeps the unobservable directions of extrinsic translation around the initial values, and outperforms the method without observability awareness. 

\section{Real-world Experiments}
\label{sec:real_experiment}
\begin{table*}[t]
\caption{The details of sequences used in real-world experiments.}
\label{tab:real_datasets}
\centering
\resizebox{0.9\linewidth}{!}{
{
    \begin{tabular}{@{}cccccccc@{}}
        \toprule
        Dataset & Sequence & Config & Duration & Env. & \begin{tabular}[c]{@{}c@{}}Ave. Angular\\ Velocity (rad/s)\end{tabular} & Motion Type & Excitation Type \\ \midrule
            & \textit{\begin{tabular}[c]{@{}c@{}}Vicon-\\ (01,02,...,10)\end{tabular}} & Hand-held & 15 & Indoor & 0.983 & Random Shaking & Full \\
            & \textit{\begin{tabular}[c]{@{}c@{}}Stairs-\\ (01,02)\end{tabular}} & Hand-held & 60 & Indoor & 1.118 & Random-Shaking & Full \\
            & \textit{\begin{tabular}[c]{@{}c@{}}Lab-\\ (01,02)\end{tabular}} & Hand-held & 60 & Indoor & 1.215 & Random Shaking & Full \\
            & \textit{SKG-01} & On-Vehicle & 120 & Outdoor & 0.407 & Planar Motion & Degenerate \\
        \multirow{-5}{*}{\begin{tabular}[c]{@{}c@{}}Self \\ collection\end{tabular}} & \textit{YQ-01} & Hand held & 271 & Outdoor & 0.760 & \begin{tabular}[c]{@{}c@{}}Casual Walking \&\\ occasional Random Shaking\end{tabular} & Occasionally full \\ \midrule
        LIO-SAM~\cite{shan2020lio} & \textit{Casual-walk} & Hand held & 655 & Outdoor & 0.521 & \begin{tabular}[c]{@{}c@{}}Casual Walking \&\\ occasional Random Shaking\end{tabular} & Occasionally full \\ \midrule
            & \textit{IC\_Office} & Hand held & 200 & Indoor & 0.348 & Casual Walking & Occasionally full \\
            & \textit{Office\_Mitte} & Hand held & 264 & Indoor & 0.350 & Casual-Walking & Occasionally full \\
        \multirow{-3}{*}{\begin{tabular}[c]{@{}c@{}}Hilti SLAM \\ challenge~\cite{helmberger2021hilti} \end{tabular}} & \textit{Basement}${{}^1}$  & On-Vehicle & 330 & Indoor & 0.116 & Planar Motion & Degenerate \\ \bottomrule
        \\[-10pt]
        \multicolumn{7}{l}{${}^1$ Corresponding to Basement\_3 sequence in Hilti dataset.}
    \end{tabular}
}
}
\end{table*}

\begin{figure}[t]
    \centering
    \includegraphics[width=1.0\columnwidth]{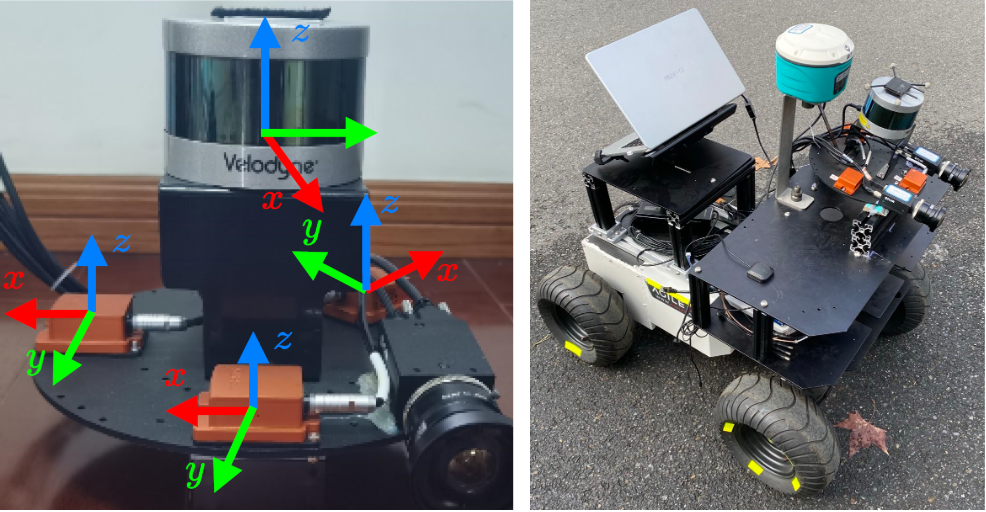}
    \caption{Left: The self-assembled sensor rig used in real-world experiments. Right: The ground wheeled robot for collecting \textit{SKG-01} sequence.}
    \label{fig:car}
\end{figure}

In the real-world experiments, we validate the proposed method on multiple our own datasets as well as two public datasets.
Table~\ref{tab:real_datasets} summarizes the sequences in terms of collection configuration, duration, environment, motion type, etc.
For the motion type:
\begin{itemize}
    \item Random Shaking:  We hold the sensor by hand and shake the sensor suite aggressively both in rotations and translations;
    \item Casual Waking: We hold the sensor and walk around;
    \item Planar Motion: The sensor rig is mounted on moving wheeled robots, see Fig.~\ref{fig:car} (right).
\end{itemize}

The self-collected datasets are collected in both indoor and outdoor scenarios by the self-assembled sensor rig shown in the left of Fig.~\ref{fig:car}, which consists of 3 Xsens-100 IMUs sampled at 400 Hz and a Velodyne VLP-16 LiDAR sampled at 10 Hz. 
The \textit{Casual-walk} sequence \footnote{Available at 
%\url{https://drive.google.com/drive/folders/1gJHwfdHCRdjP7vuT556pv8atqrCJPbUq}
\url{http://www.udel.edu/009352}}
is from LIO-SAM~\cite{shan2020lio} and is collected by a Microstrain 3DM-GX5-25 IMU sampled at 500 Hz and a Velodyne VLP-16 LiDAR sampled at 10 Hz.
The Hilti SLAM challenge dataset~\cite{helmberger2021hilti} is intended for assessing the accuracy of state estimation in challenging environments and it contains a number of visual, LiDAR and inertial sensors. In the experiments, the data from Bosch BMI085 IMU sampled at 200 Hz and the Ouster OS0-64 LiDAR sampled at 10 Hz are utilized for evaluation.

The proposed method is extensively tested on the \textit{Stairs} and \textit{Lab} sequences to examine the accuracy and repeatability of calibrated results.
Due to the absence of ground-truth extrinsic transformation between LiDAR and IMU in real-world experiments, the relative poses of three components IMU inferred from CAD assembly drawings are introduced as references. 
Additionally, the proposed observability-aware calibration method incorporating the informative data selection and observability-aware state update mechanism is thoroughly evaluated.
On the one hand, we examine the feasibility of the proposed method in degenerate motion cases over the \textit{SKG-01} sequence and \textit{Basement} sequence, which are collected on the vehicles under planar motion.
On the other hand, we also analyze the proposed informative segment selection algorithm in the \textit{YQ-01}, \textit{Casual-walk}, \textit{IC\_Office}, \textit{Office\_Mitte} sequences, among which the former two sequences are outdoors while the latter two are indoors.
Note that only one of the three IMUs (Fig.~\ref{fig:car}), IMU1, is utilized to evaluate the \textit{SKG-01} and \textit{YQ-01} sequences.

We also compare the proposed method with the open-source LiDAR-IMU calibration toolkit Imu\_Lidar\_Calibration~\cite{mishra2021target} (abbreviated as ILC) and our previous method LI-Calib~\cite{lv2020targetless}. Notably, LI-Calib~\cite{lv2020targetless} does not have the observability-aware scheme described in Sec.~\ref{sec:observability} or the explicit initialization of control points (Sec.~\ref{sec:initialization of control points}),
compared to the proposed OA-LICalib. 
In addition, two online calibration methods, LIO-Mapping~\cite{ye2019tightly} (abbreviated as LIOM) and FAST-LIO2~\cite{xu2021fast}, are also included for the comparisons. LIOM and FAST-LIO2 are LiDAR-IMU odometry and simultaneously calibrate the extrinsic parameters between LiDAR and IMU.
In practice, online calibration methods estimate the whole trajectory of the entire calibration sequence, and the final converged values of the extrinsic parameters are deemed the calibration results, as are the ILC method. Since the ILC method particularly requires the sensor to keep static for seconds to initialize the system, we skip some data segments at the beginning in order to satisfy the static initialization requirements in some sequences.
Notably, under general motion scenarios, the proposed OA-LICalib and our previous method LI-Calib are initialized from the identity extrinsic rotation matrix and zero extrinsic translations without bells and whistles. However, for the initial values of the extrinsics for these compared methods, including ILC, LIOM, and FAST-LIO2, the extrinsic rotation parameters are initialized by the reference parameters provided by the dataset, while the translation extrinsic parameters are initialized as zero. These compared methods fail to converge when starting from the identity extrinsic rotation matrix due to the lack of an extrinsic initialization module.

\begin{figure}[t]
	\centering
	\begin{subfigure}{0.49\columnwidth} 
		\includegraphics[width=\columnwidth]{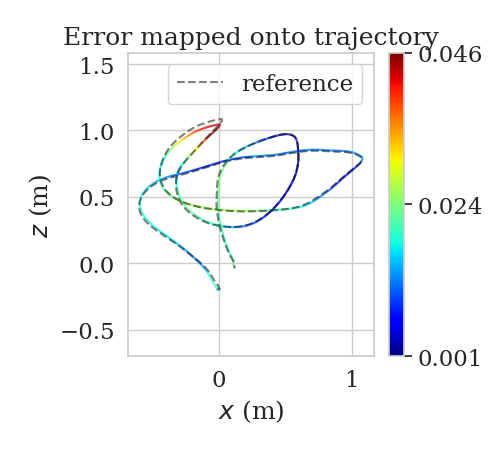}
	\end{subfigure}
	\hfill	
	\begin{subfigure}{0.49\columnwidth} 
		\includegraphics[width=\columnwidth]{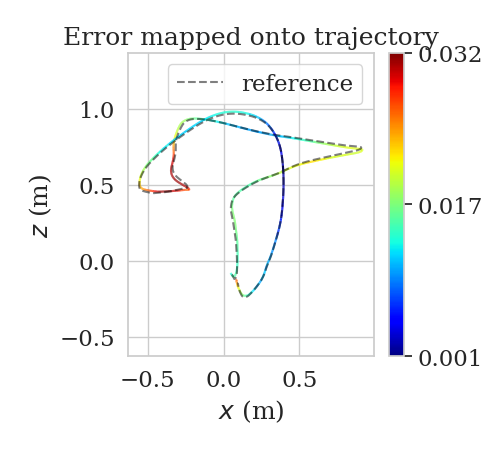}
	\end{subfigure}
	\begin{subfigure}{1.0\columnwidth} 
		\includegraphics[width=\columnwidth]{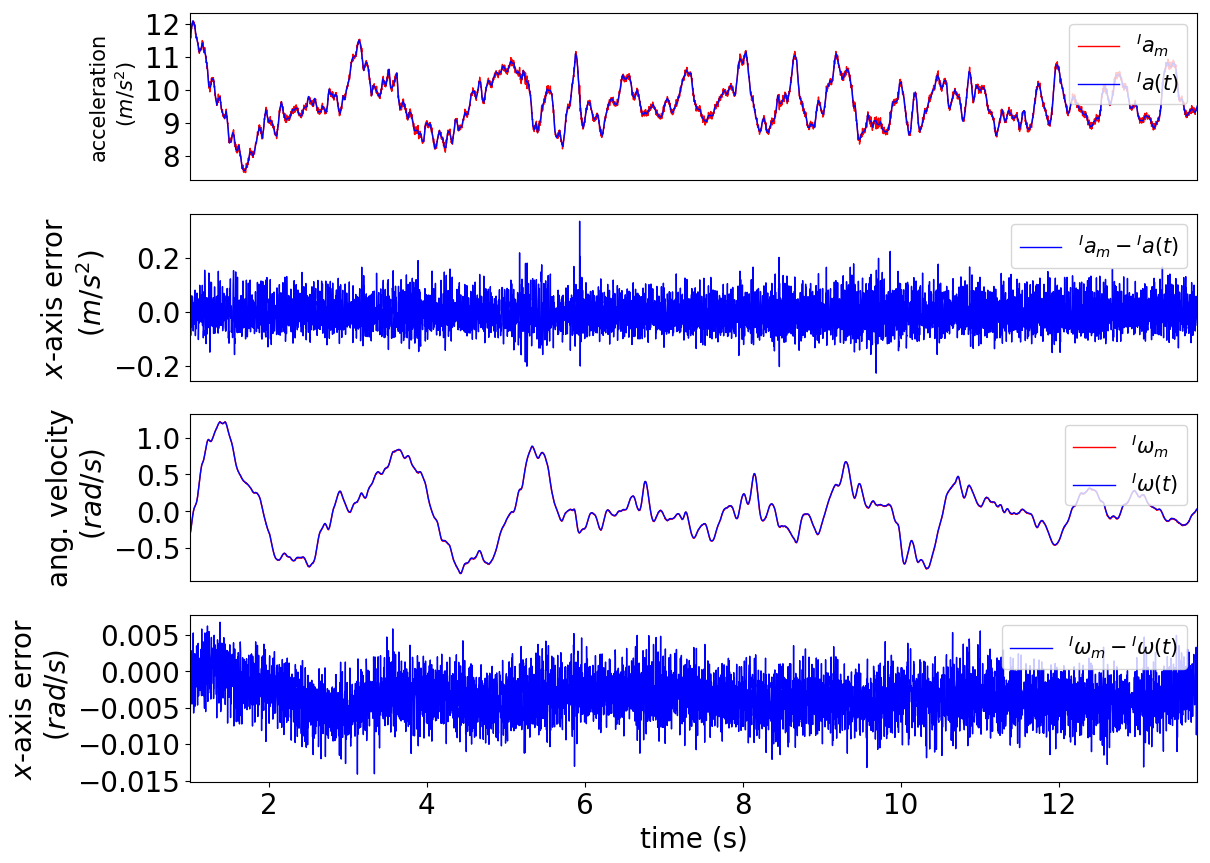}
	\end{subfigure}
	\caption{Top: Estimated continuous-time trajectories aligned with the ground-truth. The color of the trajectory indicates ATE. Bottom: The corresponding fitting results of the top-left trajectory. Only the $x$-axis is plotted. The fitting errors almost have the same order of magnitude as the IMU measurement noises.} 
	\label{fig:trajApe}
\end{figure}

\subsection{Trajectory Representation Capability}

In the \textit{Vicon} sequences, ground truth poses are provided by the Vicon system at 100 Hz, with a duration of about 15 seconds to evaluate the accuracy of estimated trajectories and the representation capability of the splines. Comparing the estimated trajectories with the motion-capture system's reported trajectories, the average absolute trajectory error (ATE) ~\cite{grupp2017evo} overall ten sequences are 0.0183 m. Figure~\ref{fig:trajApe} shows two typical estimated trajectories aligned with the ground truth and also illustrates fitting results. 
The estimated B-Spline trajectory fits well with the ground-truth trajectory and the acceleration measurements and angular velocity measurements, indicating the B-Spline's high representation capability.

\begin{figure}[t]
    \centering
	\includegraphics[width=\columnwidth]{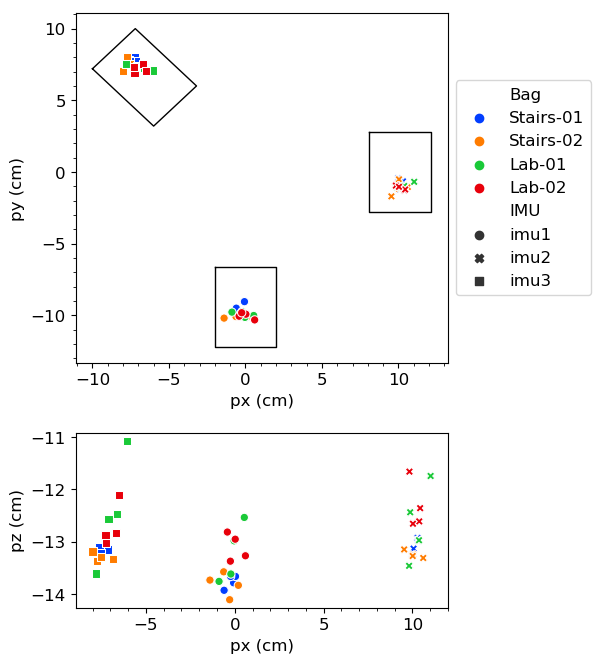}
	\caption{Extrinisc translation calibration results from OA-LICalib over \textit{Stairs} and \textit{Lab} sequences. Top: Top view of the calibration results. The three black rectangular boxes represent the actual mounting positions of three IMUs, respectively. Bottom: Front view of the calibration results.}
	\label{fig:6dof-position}
\end{figure}

\begin{figure}[t]
	\centering
	\includegraphics[width=\columnwidth]{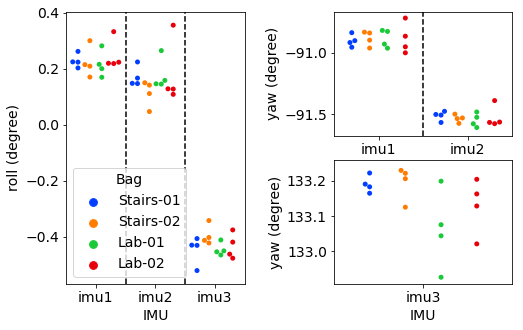}
	\caption{Extrinsic Rotation calibration results from OA-LICalib over \textit{Stairs} and \textit{Lab} sequences.}
	\label{fig:6dof-rotation}
\end{figure}

\begin{table*}[t]
\caption{The mean and standard deviation of the calibration results from different methods on \textit{Stairs} and \textit{Lab} sequences. 
The units for translation and rotation are in `cm' and `degree', respectively.
The non-converged results of ILC and LIOM are excluded from the statistics.
}
% \resizebox{\linewidth}{!}{
\centering
{
\begin{tabular}{|c|c|r@{}c@{}lr@{}c@{}lr@{}c@{}lr@{}c@{}lr@{}c@{}lr@{}c@{}l|cc|}
    \hline
    &  & \multicolumn{9}{c|}{\cellcolor[HTML]{E0EFD4}${}^L\bp_I$ (cm)} & \multicolumn{9}{c|}{\cellcolor[HTML]{E0EFD4}${}^L_I\bar{q}$ in Euler (deg)} & \multicolumn{2}{c|}{\cellcolor[HTML]{EECC89}RMSE} \\[1pt] \cline{3-22} 
   \multirow{-2}{*}{Method} & \multirow{-2}{*}{IMU} & \multicolumn{3}{c}{\cellcolor[HTML]{E0EFD4}$p_x$} & \multicolumn{3}{c}{\cellcolor[HTML]{E0EFD4}$p_y$} & \multicolumn{3}{c|}{\cellcolor[HTML]{E0EFD4}$p_z$} & \multicolumn{3}{c}{\cellcolor[HTML]{E0EFD4}roll} & \multicolumn{3}{c}{\cellcolor[HTML]{E0EFD4}pitch} & \multicolumn{3}{c|}{\cellcolor[HTML]{E0EFD4}yaw} & \cellcolor[HTML]{EECC89}Pos (cm) & \cellcolor[HTML]{EECC89}Rot (deg) \\ \hline
    & IMU1 & \multicolumn{3}{c}{0} & \multicolumn{3}{c}{-10} & \multicolumn{3}{c}{-13} & \multicolumn{3}{c}{0} & \multicolumn{3}{c}{0} & \multicolumn{3}{c|}{-90} & - & - \\
    & IMU2 & \multicolumn{3}{c}{10} & \multicolumn{3}{c}{0} & \multicolumn{3}{c}{-13} & \multicolumn{3}{c}{0} & \multicolumn{3}{c}{0} & \multicolumn{3}{c|}{-90} & - & - \\
   \multirow{-3}{*}{CAD Reference} & IMU3 & \multicolumn{3}{c}{-7} & \multicolumn{3}{c}{8} & \multicolumn{3}{c}{-13} & \multicolumn{3}{c}{0} & \multicolumn{3}{c}{0} & \multicolumn{3}{c|}{135} & - & - \\ \hline
    & IMU1 & 9.61 & $\pm$ & 0.79 & -13.00 & $\pm$ & 4.78 & -7.34 & $\pm$ & 6.75 & 7.78 & $\pm$ & 0.47 & -4.95 & $\pm$ & 5.83 & -95.82 & $\pm$ & 2.18 & 11.65 & 5.04 \\
    & IMU2 & 19.65 & $\pm$ & 5.63 & -1.69 & $\pm$ & 3.36 & -4.44 & $\pm$ & 3.54 & 9.73 & $\pm$ & 1.40 & -3.48 & $\pm$ & 3.35 & -93.59 & $\pm$ & 1.59 & 6.67 & 6.30 \\
   \multirow{-3}{*}{ILC} & IMU3 & 7.74 & $\pm$ & 2.13 & 7.20 & $\pm$ & 2.33 & -10.95 & $\pm$ & \textbf{0.49} & -6.63 & $\pm$ & 2.04 & 7.69 & $\pm$ & 2.05 & 127.47 & $\pm$ & 1.35 & 7.51 & 6.32 \\ \hline
    & IMU1 & -0.74 & $\pm$ & 3.77 & -12.87 & $\pm$ & 2.40 & -2.91 & $\pm$ & 4.31 & 9.64 & $\pm$ & 1.19 & 0.09 & $\pm$ & 0.25 & -93.31 & $\pm$ & 2.37 & 8.60 & 7.30 \\
    & IMU2 & 7.70 & $\pm$ & 6.42 & -5.85 & $\pm$ & 2.63 & -2.68 & $\pm$ & 2.71 & 9.82 & $\pm$ & 1.71 & 0.19 & $\pm$ & 0.29 & -94.06 & $\pm$ & 2.50 & 6.98 & 6.13 \\
   \multirow{-3}{*}{LIOM} & IMU3 & -2.64 & $\pm$ & 2.54 & 11.21 & $\pm$ & 7.15 & 6.43 & $\pm$ & 3.96 & 7.89 & $\pm$ & 4.09 & 0.32 & $\pm$ & 0.27 & 131.30 & $\pm$ & 2.97 & 11.65 & 5.04 \\ \hline
    & IMU1 & -1.56 & $\pm$ & 1.28 & -10.56 & $\pm$ & 3.24 & -12.94 & $\pm$ & 2.78 & -0.52 & $\pm$ & 1.65 & 1.49 & $\pm$ & 1.06 & -91.00 & $\pm$ & 0.92 & 0.96 & 1.08 \\
    & IMU2 & 6.76 & $\pm$ & 1.89 & -3.08 & $\pm$ & 2.40 & -12.39 & $\pm$ & 1.96 & -1.05 & $\pm$ & 2.00 & 1.45 & $\pm$ & 0.89 & -91.34 & $\pm$ & 1.32 & 2.61 & 1.29 \\
   \multirow{-3}{*}{FAST-LIO2} & IMU3 & -7.84 & $\pm$ & 1.58 & 4.48 & $\pm$ & 2.72 & -10.41 & $\pm$ & 0.85 & -0.34 & $\pm$ & 1.03 & -0.7 & $\pm$ & 2.13 & 132.96 & $\pm$ & 0.78 & 2.57 & 1.26 \\ \hline
    & IMU1 & -0.97 & $\pm$ & 2.37 & -9.80 & $\pm$ & 1.49 & -12.98 & $\pm$ & 1.41 & 0.47 & $\pm$ & 0.40 & 0.18 & $\pm$ & 0.87 & -90.90 & $\pm$ & 0.29 & 0.57 & 0.60 \\
    & IMU2 & 9.64 & $\pm$ & 2.09 & -1.08 & $\pm$ & 1.10 & -12.47 & $\pm$ & 1.25 & 0.36 & $\pm$ & 0.39 & 0.42 & $\pm$ & 0.83 & -91.56 & $\pm$ & 0.27 & 0.73 & 0.96 \\
   \multirow{-3}{*}{LI-Calib} & IMU3 & -6.73 & $\pm$ & 2.03 & 7.09 & $\pm$ & 1.33 & -12.66 & $\pm$ & 1.11 & -0.56 & $\pm$ & 0.60 & 0.89 & $\pm$ & 0.81 & 133.14 & $\pm$ & 0.31 & 0.58 & 1.23 \\ \hline
    & IMU1 & -0.22 & $\pm$ & \textbf{0.50} & -9.92 & $\pm$ & \textbf{0.31} & -13.48 & $\pm$ & \textbf{0.44} & 0.23 & $\pm$ & \textbf{0.04} & 0.21 & $\pm$ & \textbf{0.06} & -90.89 & $\pm$ & \textbf{0.07} & \textbf{0.31} & \textbf{0.54} \\
    & IMU2 & 10.16 & $\pm$ & \textbf{0.36} & -1.00 & $\pm$ & \textbf{0.31} & -12.83 & $\pm$ & \textbf{0.54} & 0.16 & $\pm$ & \textbf{0.07} & 0.46 & $\pm$ & \textbf{0.17} & -91.53 & $\pm$ & \textbf{0.05} & \textbf{0.59} & \textbf{0.93} \\
   \multirow{-3}{*}{OA-LICalib} & IMU3 & -7.19 & $\pm$ & \textbf{0.53} & 7.46 & $\pm$ & \textbf{0.34} & -12.89 & $\pm$ & 0.61 & -0.43 & $\pm$ & \textbf{0.04} & 0.68 & $\pm$ & \textbf{0.05} & 133.14 & $\pm$ & \textbf{0.09} & \textbf{0.33} & \textbf{1.17} \\ \hline
   \end{tabular}
}
\label{tab:real_full_motion_compare}
\end{table*}

\begin{table}[t]
\caption{The OA-LICalib's time consumption of main stages. Two  segments from \textit{Lab-01} and \textit{SKG-01} sequence are evaluated, respectively. The former segment is under fully-excited motion, while the latter is under degeneration motion.}
\centering
 {
\resizebox{\linewidth}{!}{
\begin{tabular}{@{}ccccccc@{}}
\toprule
\multirow{2}{*}{Step} & \multicolumn{3}{c}{\begin{tabular}[c]{@{}c@{}}Fully-excited Seq.\\ {[}0,15{]} of \textit{Lab-01}\end{tabular}} & \multicolumn{3}{c}{\begin{tabular}[c]{@{}c@{}}Degenerate Seq.\\ {[}0,15{]} of \textit{SKG-01}\end{tabular}} \\ \cmidrule(l){2-7} 
 & Count & \begin{tabular}[c]{@{}c@{}}Processing\\ /Step(s)\end{tabular} & Total(s) & Count & \begin{tabular}[c]{@{}c@{}}Processing\\ /Step(s)\end{tabular} & Total(s) \\ \midrule
Extri. Init. \& Rerun Odom. & 1 & 15.73 & 15.73 & 1 & 12.25 & 12.25 \\
Data Association & 1 & 1.80 & 1.80 & 1 & 3.31 & 3.31 \\
Init. Traj. \&  First Opt. & 1 & 4.47 & 4.47 & 1 & 6.54 & 6.54 \\
Refinement & 5 & 6.73 & 33.64 & 5 & 8.03 & 40.17 \\
TSVD & 0 & - & - & 5 & 0.03 & 0.13 \\ \midrule
Total & - & - & 55.65 & - & - & 64.40 \\ \bottomrule
\end{tabular}
}
}
\label{tab:time_consumption}
\end{table}

\subsection{Calibration with Full Motion Excitation}

We gathered four sequences of a 60-second duration with full excitation in structured environments, \textit{Stairs-01 and Stairs-02} collected at the stairway, \textit{Lab-01 and Lab-02} in the laboratory, verifying the calibration accuracy of the full motion excitation data. Each sequence is split into four segments of 15 seconds and calibrated independently for calibration test. The extrinsic calibration results of three IMUs with respect to the LiDAR by the proposed OA-LICalib are shown in Fig.~\ref{fig:6dof-position} and Fig.~\ref{fig:6dof-rotation}, where the black rectangles indicate the CAD sketch of three IMUs. 
We can intuitively find that the translational calibration results are distributed within rectangles, with the distribution range in translation at around 2cm. The distribution range of rotation concentrates within 0.2 degrees for each IMU.
These results demonstrate the proposed method's high repeatability and reliability in practical deployments. 

Additionally, the quantitative comparison results are summarized in Table~\ref{tab:real_full_motion_compare}, where RMSE is calculated by comparing it to the CAD reference. Note that LIOM fails to converge on the \textit{Stairs} sequences and ILC does not converge on the \textit{Lab} sequences in our experiments. Failing to convergence means rotational RMSE is greater than 30 degrees. The non-converged results are excluded from the statistics in Table~\ref{tab:real_full_motion_compare}.
Although the discrete-time methods, ILC, LIOM and FAST-LIO are initialized from the reference extrinsic rotation, the continuous-time based methods LI-Calib and OA-LICalib show apparent advantages over all of them, while starting from the identity extrinsic rotation matrix.
In addition, compared to LI-Calib, OA-LICalib takes more effort on the initialization of control points (Sec.~\ref{sec:initialization of control points}), which benefits the convergence and improves the accuracy of calibrations. It is worthwhile to mention that the calibration results of OA-LICalib are with high calibration accuracy yet low standard deviations, which demonstrate its great reliability.  
To give an impression of the efficiency, the proposed OA-LICalib's time consumption of main stages in two typical 15s segment over the sequence \textit{Lab-01} (under fully-excited motion) and sequence \textit{SKG-01} (under degenerate motion) are summarized in Table~\ref{tab:time_consumption}. Basically, the proposed method takes about only one minute to generate the highly-accurate calibration results automatically on a 15s calibration segment.

\begin{figure}[t]
    \begin{subfigure}{1.0\columnwidth} 
	    \centering
		\includegraphics[width=1.0\columnwidth]{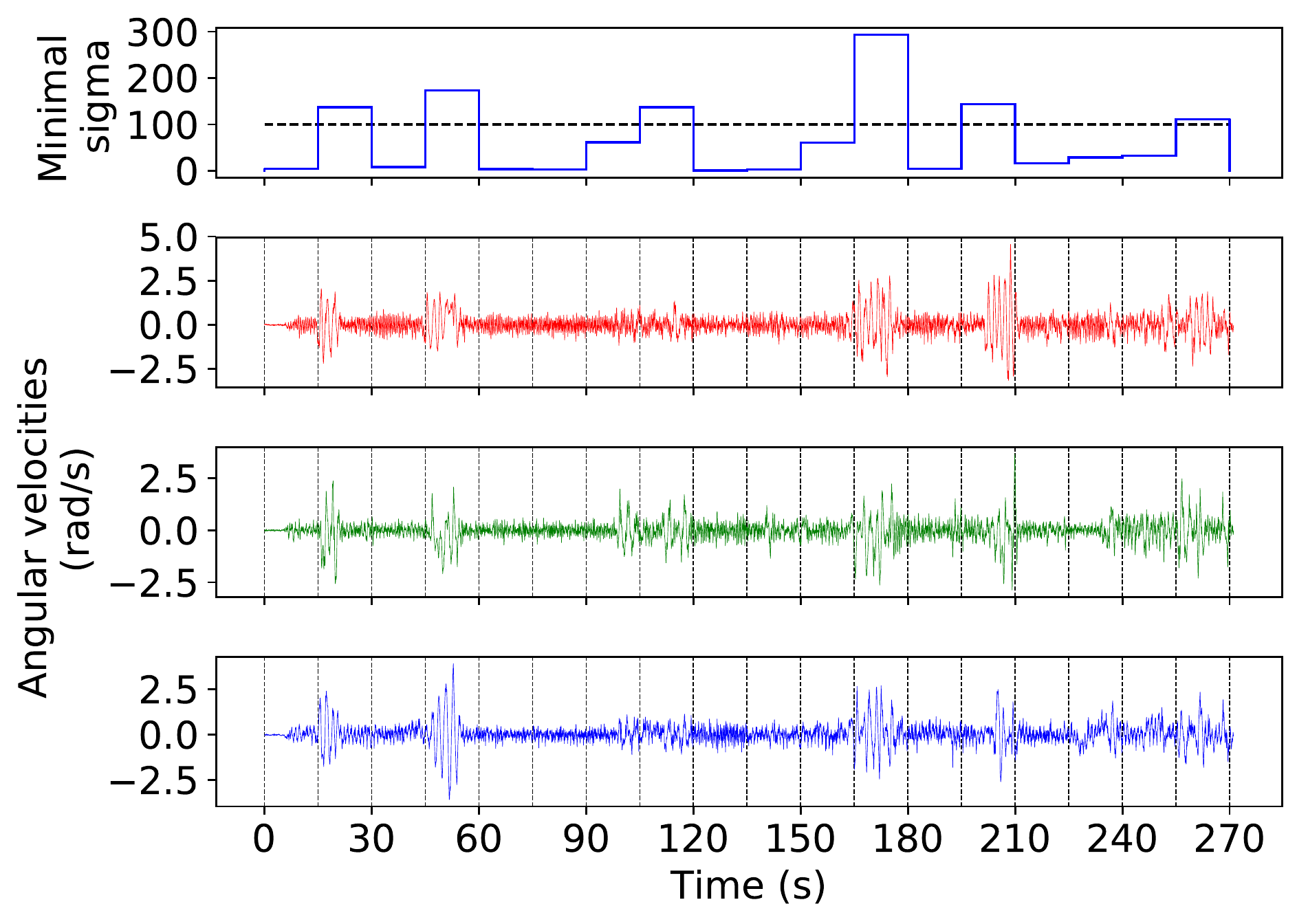}
    \end{subfigure}
	\begin{subfigure}{1.0\columnwidth} 
	    \centering
		\includegraphics[width=1.0\columnwidth]{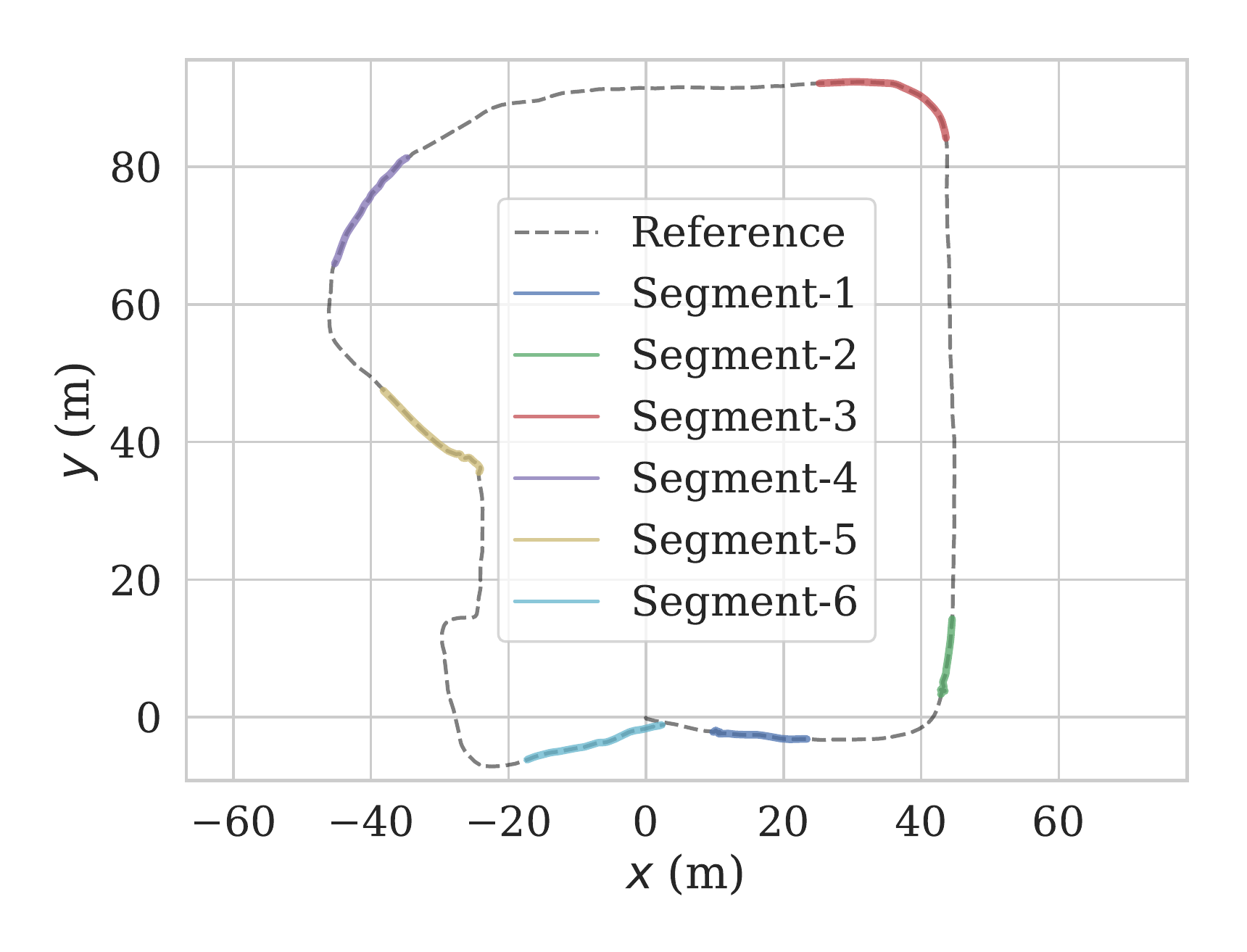}
    \end{subfigure}
	\caption{Evaluation in \textit{YQ-01} sequence. Top: The informativeness of every individual segment of \textit{YQ-01} sequence, which has a total trajectory length of 363 meters; the corresponding three-axis angular velocity curves. Bottom: The whole trajectory is plotted in dashed line and the selected segments in bright colors. The trajectory direction is counterclockwise. }
	\label{fig:segment_skg}
\end{figure}

\begin{table}[t]
\caption{The estimated spatial parameters $\{{}^I_L\bar{q},{}^I\bp_{L}\}$ and time offset $t_c$ using one segment (Segment-$n$) or multi-segment ($n$ Segments) over \textit{YQ-01} sequence. 
} 
\label{table:segment_skg}
\centering
\resizebox{\linewidth}{!}
{
\begin{tabular}{|c|ccccccc|cc|}
\hline
 & \multicolumn{3}{c|}{\cellcolor[HTML]{E0EFD4}${}^I\bp_{L}$ (cm)} & \multicolumn{3}{c|}{\cellcolor[HTML]{E0EFD4}${}^I_L\bar{q}$ in Euler (deg)} & \cellcolor[HTML]{E0EFD4} & \multicolumn{2}{c|}{\cellcolor[HTML]{EECC89}RMSE} \\ \cline{2-7} \cline{9-10} 
\multirow{-2}{*}{Data} & \cellcolor[HTML]{E0EFD4}$p_x$ & \cellcolor[HTML]{E0EFD4}$p_y$ & \multicolumn{1}{c|}{\cellcolor[HTML]{E0EFD4}$p_z$} & \cellcolor[HTML]{E0EFD4}roll & \cellcolor[HTML]{E0EFD4}pitch & \multicolumn{1}{c|}{\cellcolor[HTML]{E0EFD4}yaw} & \multirow{-2}{*}{\cellcolor[HTML]{E0EFD4}\begin{tabular}[c]{@{}c@{}}Time\\Offset (ms)\end{tabular}} & \cellcolor[HTML]{EECC89}Pos (cm) & \cellcolor[HTML]{EECC89}Rot (deg) \\ \hline
CAD Reference & -10 & 0 & 13 & 180 & -180 & -90 & - &  &  \\ \hline
Segment-1 & -11.0 & 0.3 & 14.6 & 179.75 & -179.76 & -89.25 & 7.4 & 1.10 & 0.48 \\
Segment-2 & -11.0 & -0.2 & 14.0 & 179.77 & -179.81 & -89.23 & 7.6 & 0.82 & \textbf{0.48} \\
Segment-3 & -11.2 & 0.7 & 13.5 & 179.77 & -179.8 & -89.18 & 7.4 & 0.85 & 0.51 \\
Segment-4 & -11.0 & 0.0 & 14.7 & 179.73 & -179.77 & -89.2 & 7.8 & 1.14 & 0.51 \\
Segment-5 & -9.7 & 0.3 & 12.8 & 179.71 & -179.81 & -89.18 & 7.4 & \textbf{0.27} & 0.51 \\
Segment-6 & -10.6 & -0.2 & 15.1 & 179.78 & -179.82 & -89.19 & 7.9 & 1.27 & 0.50 \\ \hline
2 Segments & -11.3 & 0.2 & 14.4 & 179.76 & -179.77 & -89.25 & - & 1.11 & \textbf{0.47} \\
3 Segments & -11.3 & 0.3 & 14.2 & 179.76 & -179.77 & -89.24 & - & 1.04 & 0.48 \\
4 Segments & -11.2 & 0.2 & 14.3 & 179.75 & -179.77 & -89.23 & - & 1.03 & 0.49 \\
5 Segments & -11.1 & 0.2 & 14.1 & 179.75 & -179.79 & -89.23 & - & \textbf{0.91} & 0.48 \\
6 Segments & -11.1 & 0.2 & 14.3 & 179.76 & -179.8 & -89.23 & - & 0.99 & 0.48 \\ \hline
\end{tabular}
}
\vspace{-1em}
\end{table}

\begin{figure}[t]
	\centering
    \includegraphics[width=1.0\columnwidth]{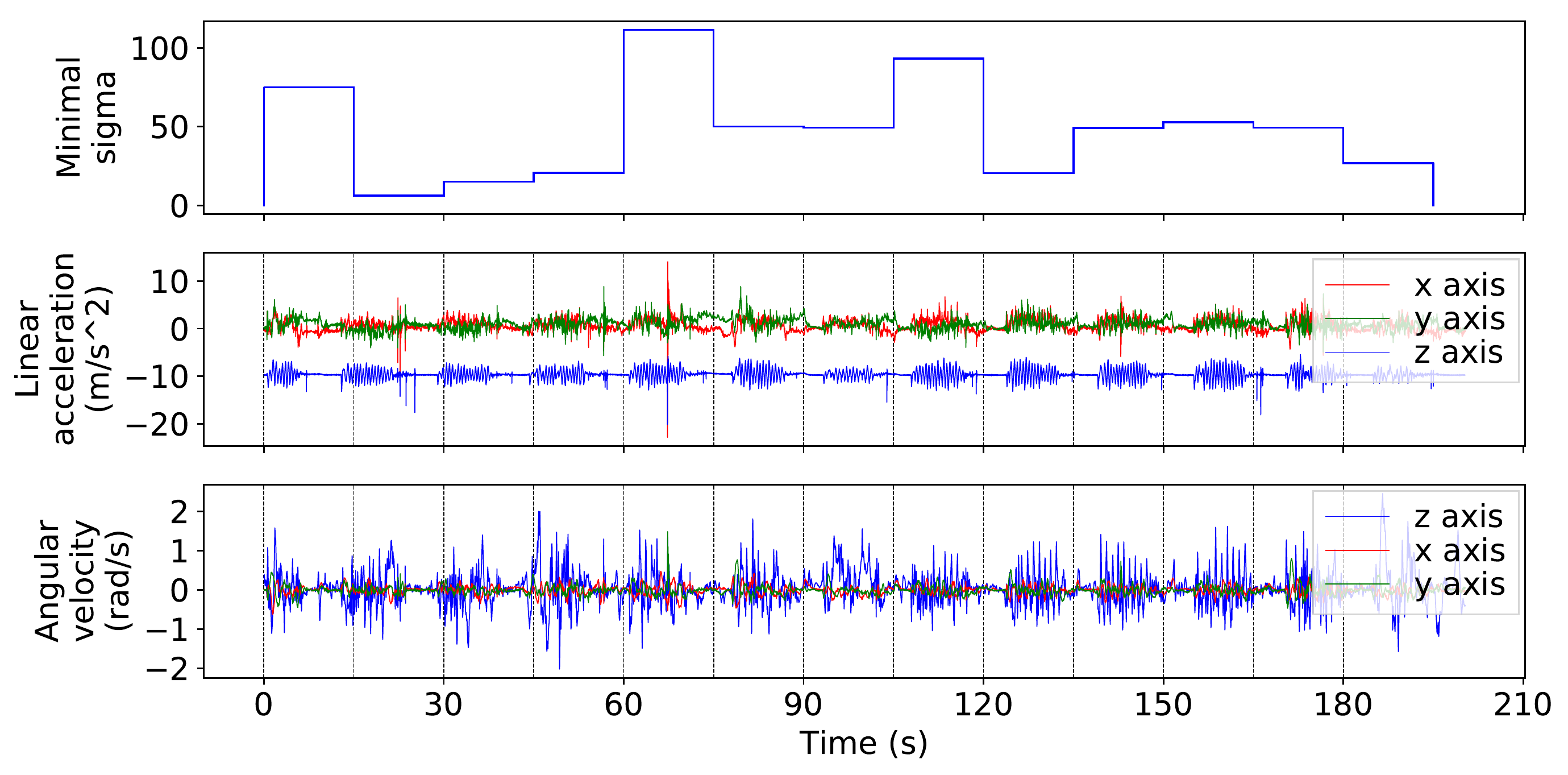}
	\caption{Top: The informativeness of every individual segment of \textit{IC-Office} sequence. Bottom: the corresponding linear acceleration curves and angular velocity curves.}
	\label{fig:hilti_ic_office}
\end{figure}

\begin{table}[h]

\caption{Comparison results of different methods on Hilti dataset. LIOM fails to give reasonable results, thus not listed in the table.
}
\centering
 {
\resizebox{\linewidth}{!} {
\begin{tabular}{|cc|cccccccc|cc|}
\hline
 &  & \multicolumn{1}{c|}{} & \multicolumn{3}{c|}{\cellcolor[HTML]{E0EFD4}} & \multicolumn{3}{c|}{\cellcolor[HTML]{E0EFD4}} & \cellcolor[HTML]{E0EFD4} & \multicolumn{2}{c|}{\cellcolor[HTML]{EECC89}} \\
 &  & \multicolumn{1}{c|}{} & \multicolumn{3}{c|}{\multirow{-2}{*}{\cellcolor[HTML]{E0EFD4}${}^I\bp_L$ (cm)}} & \multicolumn{3}{c|}{\multirow{-2}{*}{\cellcolor[HTML]{E0EFD4}${}^I_L\bar{q}$ in Euler (deg)}} & \cellcolor[HTML]{E0EFD4} & \multicolumn{2}{c|}{\multirow{-2}{*}{\cellcolor[HTML]{EECC89}RMSE}} \\ \cline{4-9} \cline{11-12} 
\multirow{-3}{*}{Method} & \multirow{-3}{*}{Sequence} & \multicolumn{1}{c|}{\multirow{-3}{*}{Data}} & \cellcolor[HTML]{E0EFD4}$p_x$ & \cellcolor[HTML]{E0EFD4}$p_y$ & \multicolumn{1}{c|}{\cellcolor[HTML]{E0EFD4}$p_z$} & \cellcolor[HTML]{E0EFD4}roll & \cellcolor[HTML]{E0EFD4}pitch & \multicolumn{1}{c|}{\cellcolor[HTML]{E0EFD4}yaw} & \multirow{-3}{*}{\cellcolor[HTML]{E0EFD4}\begin{tabular}[c]{@{}c@{}}Time\\ Offset\\ (ms)\end{tabular}} & \cellcolor[HTML]{EECC89}Pos (cm) & \cellcolor[HTML]{EECC89}Rot (deg) \\ \hline
\multicolumn{2}{|c|}{Reference} &  & 1.00 & -0.66 & 9.47 & 0.02 & -0.37 & -0.14 & $\sim$1ms &  &  \\ \hline
\multicolumn{1}{|c|}{} & \textit{Office\_Mitte} & All data & 0.43 & -1.02 & 6.58 & 0.05 & -0.02 & -0.35 & - & 1.71 & 0.24 \\ \cline{2-12} 
\multicolumn{1}{|c|}{\multirow{-2}{*}{FAST-LIO2}} & IC\_Office & All data & 0.69 & -1.21 & 5.02 & 0.03 & 0.12 & -0.59 & - & 2.59 & 0.39 \\ \hline
\multicolumn{1}{|c|}{} & \textit{Office\_Mitte} & All data & -0.02 & -1.09 & 11.95 & 1.54 & -0.50 & 0.12 & -3.86 & 1.57 & 0.89 \\ \cline{2-12} 
\multicolumn{1}{|c|}{\multirow{-2}{*}{ILC}} & IC\_Office & All data & -0.24 & 3.65 & 10.95 & -0.01 & 0.97 & 5.16 & -11.90 & 2.73 & 3.15 \\ \hline
\multicolumn{1}{|c|}{} &  & Segment-1 & 0.20 & -0.70 & 9.10 & 0.12 & -0.07 & -0.26 & 1.62 & \textbf{0.51} & 0.20 \\
\multicolumn{1}{|c|}{} &  & Segment-2 & 0.71 & 0.86 & 5.57 & 0.25 & -0.31 & 0.45 & 1.76 & 2.42 & 0.37 \\
\multicolumn{1}{|c|}{} &  & Segment-3 & -0.44 & -0.66 & 6.96 & 0.04 & -0.20 & -0.08 & 1.65 & 1.67 & \textbf{0.10} \\
\multicolumn{1}{|c|}{} &  & Segment-4 & -2.64 & -1.74 & 9.73 & 0.00 & 0.10 & 0.09 & 1.53 & 2.20 & 0.30 \\ \cline{3-12} 
\multicolumn{1}{|c|}{} &  & 2-Segment & -0.15 & 0.16 & 7.21 & 0.16 & -0.14 & -0.11 & - & 1.54 & 0.16 \\
\multicolumn{1}{|c|}{} &  & 3-Segment & -0.22 & 0.25 & 7.72 & 0.12 & -0.15 & -0.15 & - & 1.34 & \textbf{0.14} \\
\multicolumn{1}{|c|}{} & \multirow{-7}{*}{\textit{Office\_Mitte}} & 4-Segment & -0.20 & 0.02 & 7.69 & 0.11 & -0.13 & -0.18 & - & \textbf{1.30} & 0.15 \\ \cline{2-12} 
\multicolumn{1}{|c|}{} &  & Segment-1 & 1.16 & -2.30 & 6.33 & -0.05 & -0.20 & -0.43 & 1.55 & 2.05 & 0.20 \\
\multicolumn{1}{|c|}{} &  & Segment-2 & -1.49 & 0.58 & 8.81 & -0.03 & -0.42 & -0.02 & 0.96 & 1.65 & 0.08 \\
\multicolumn{1}{|c|}{} &  & Segment-3 & 0.77 & 2.17 & 5.67 & 0.10 & -0.31 & -0.09 & 1.30 & 2.74 & \textbf{0.06} \\
\multicolumn{1}{|c|}{} &  & Segment-4 & -0.17 & 0.17 & 7.34 & -0.20 & -0.25 & -0.07 & 1.43 & \textbf{1.48} & 0.15 \\ \cline{3-12} 
\multicolumn{1}{|c|}{} &  & 2-Segment & 0.48 & -0.86 & 7.37 & -0.03 & -0.27 & -0.45 & - & \textbf{1.25} & 0.19 \\
\multicolumn{1}{|c|}{} &  & 3-Segment & 0.64 & 0.24 & 6.76 & 0.01 & -0.27 & -0.27 & - & 1.66 & 0.09 \\
\multicolumn{1}{|c|}{\multirow{-14}{*}{OA-LICalib}} & \multirow{-7}{*}{\textit{IC\_Office}} & 4-Segment & 0.53 & 0.30 & 6.91 & -0.02 & -0.27 & -0.23 & - & 1.60 & \textbf{0.08} \\ \hline
\end{tabular}
}
}
\label{tab:hilti_multi_segment}
\end{table}

\begin{table}[t]
\caption{Multiple-segment calibration using non-informative segment in \textit{IC-Office} sequences. The segments \colorbox[HTML]{C4FFC4}{[60,75]} and \colorbox[HTML]{C4FFC4}{[105,120]} are identified as informative segments and segment \colorbox[HTML]{FFCCC9}{[15,30]} is identified as non-informative segment, as shown in Fig.~\ref{fig:hilti_ic_office}. }
\centering
 {
\resizebox{\linewidth}{!}{
\begin{tabular}{|c|ccccccc|cc|}
\hline
 & \multicolumn{3}{c|}{\cellcolor[HTML]{E0EFD4}} & \multicolumn{3}{c|}{\cellcolor[HTML]{E0EFD4}} & \cellcolor[HTML]{E0EFD4} & \multicolumn{2}{c|}{\cellcolor[HTML]{EECC89}} \\
 & \multicolumn{3}{c|}{\multirow{-2}{*}{\cellcolor[HTML]{E0EFD4}${}^I\bp_L$ (cm)}} & \multicolumn{3}{c|}{\multirow{-2}{*}{\cellcolor[HTML]{E0EFD4}${}^I_L\bar{q}$ in Euler (deg)}} & \cellcolor[HTML]{E0EFD4} & \multicolumn{2}{c|}{\multirow{-2}{*}{\cellcolor[HTML]{EECC89}RMSE}} \\ \cline{2-7} \cline{9-10} 
\multirow{-3}{*}{Data} & \cellcolor[HTML]{E0EFD4}$p_x$ & \cellcolor[HTML]{E0EFD4}$p_y$ & \multicolumn{1}{c|}{\cellcolor[HTML]{E0EFD4}$p_z$} & \cellcolor[HTML]{E0EFD4}roll & \cellcolor[HTML]{E0EFD4}pitch & \multicolumn{1}{c|}{\cellcolor[HTML]{E0EFD4}yaw} & \multirow{-3}{*}{\cellcolor[HTML]{E0EFD4}\begin{tabular}[c]{@{}c@{}}Time \\ Offset\\ (ms)\end{tabular}} & \cellcolor[HTML]{EECC89}Pos (cm) & \cellcolor[HTML]{EECC89}Rot (deg) \\ \hline
Reference & 1.00 & -0.66 & 9.47 & 0.02 & -0.37 & -0.14 & $\sim$1ms &  &  \\ \hline
\cellcolor[HTML]{C4FFC4}{[}60,75{]} & 1.16 & -2.30 & 6.33 & -0.05 & -0.20 & -0.43 & 1.55 & 2.05 & 0.20 \\
\cellcolor[HTML]{C4FFC4}{[}105,120{]} & -1.49 & 0.58 & 8.81 & -0.03 & -0.42 & -0.02 & 0.96 & 1.65 & \textbf{0.08} \\
\cellcolor[HTML]{FFCCC9}{[}15,30{]} & 0.68 & -2.84 & 1.47 & 0.09 & -0.45 & 0.01 & 0.66 & 4.79 & 0.11 \\ \hline
{[}60,75{]}+{[}105,120{]} & 0.48 & -0.86 & 7.37 & -0.03 & -0.27 & -0.45 & - & \textbf{1.25} & 0.19 \\
{[}60,75{]}+{[}15,30{]} & 0.69 & -2.75 & 3.91 & -0.01 & -0.33 & -0.42 & - & 3.43 & 0.16 \\ \hline
\end{tabular}
}
}
\label{tab:hilti_non_informative_segment}
\vspace{-1em}
\end{table}

\subsection{Calibration with Informative Segments} 
There is a relatively common situation in practice where some datasets provide a large number of various data sequences yet without accurate extrinsic parameters. There's no possibility of collecting new sequences with the sensors used in the existing datasets. Hence, if users want to get more accurate extrinsic parameters, they need to complete the calibration based on the existing data sequences. Special care should be taken when identifying informative sequences for accurate calibration.

Regarding the implementation details about multi-segment calibration, we firstly perform single segment calibration and identify informative segments;
afterward, with all the measurements from selected informative segments, we conduct a total continuous-time batch optimization to simultaneously estimate each segment's individual states (including the continuous-time trajectory, time offset $t_c$, IMU navigation states $\bx_{Is}$), and the shared spatial extrinsic parameters $\{{}^I_L\bar{q},{}^I\bp_{L}\}$.
We evaluate the proposed informativeness-aware segment selection algorithm in the self-collected sequence \textit{YQ-01} and public-available sequence \textit{Casual-walk} \textit{and Office\_Mitte}, \textit{IC\_Office} from the Hilti dataset. Like the \textit{Casual-walk} sequence collection process, we walk in a circle profile with the occasional shake of the sensor to collect \textit{YQ-01} sequence. Again, we split sequences into segments with a duration of 15-second for each segment. 

\subsubsection{\textit{YQ-01} Sequence}
Figure~\ref{fig:segment_skg} details the shape of the trajectory and the corresponding angular velocity curves of the \textit{YQ-01} sequence.
The top figure in Fig.~\ref{fig:segment_skg} illustrates each segment's information metrics and angular velocities.
The segments with minimal singular value (see Sec.~\ref{sec:data segment selection}) above 100 (black dash line) are selected as the informative segments and the relative location of these segments are shown in bright colors at the bottom of Fig.~\ref{fig:segment_skg}. It is interesting to see the amplitudes of angular velocity can indistinctly reflect the information metrics.
The extrinsic calibration results with data of one segment (Segment-$n$) and multi-segment ($n$ Segments) are summarized in Table~\ref{table:segment_skg}.
Note that ``$n$ Segments" means we take the first $n$ informative segments, e.g., ``3 Segments" represents Segment-1, Segment-2 and Segment-3 are jointly utilized for calibration.
The evaluation results show that the selected single-segment calibration is able to generate reasonable calibration results, which demonstrate the effectiveness of the informative segment selection.
Besides, multi-segment calibration is generally stable compared to single-segment calibration, and four informative segments are enough to get a reliable estimation of the extrinsics.

\subsubsection{Hilti Sequences}
Furthermore, we test the multi-segment calibration on two handheld sequences of the Hilti dataset, the comparison results are reported in Table~\ref{tab:hilti_multi_segment}. 
Figure~\ref{fig:hilti_ic_office} reports the information metrics of every individual segment, linear acceleration curves, and angular velocity curves in the \textit{IC\_Office} sequence, and we pick out the four most informative segments for multi-segment calibration. The results in Table~\ref{tab:hilti_multi_segment} suggest multi-segment calibration is more stable and more accurate than single-segment calibration. The proposed OA-LICalib with multi-segment calibration also significantly outperforms ILC and FAST-LIO2.

Note that using minimal singular value to measure the informativeness of each individual segment within a single sequence is feasible, however it is not meaningful to compare the minimal singular values of different sequences under significantly different scenarios.
Minimal singular value is not a general information metric~\cite{schneider2019observability}, and it is affected by the sensory observations, trajectory profile, surrounding environment, etc. This is reflected by the different amplitudes of minimal singular values over different sequences, as shown in Fig.~\ref{fig:segment_skg} and Fig.~\ref{fig:hilti_ic_office}. 
On the positive side, a minimal-singular-value-based information metric is sufficient for informative segments selection from a given sequence and leads to reasonable calibration results.

We further investigate the performance of non-informative segments and their effect on multi-segment calibration results. Table~\ref{tab:hilti_non_informative_segment} shows the calibration results using multiple segments in \textit{IC-Office} sequences. The segments [60, 75] and [105, 120] are identified as informative segments, while the segment [15, 30] is a non-informative segment automatically by OA-LICalib, as shown in Fig.~\ref{fig:hilti_ic_office}. The results in Table~\ref{tab:hilti_non_informative_segment} suggest that the non-informative segment generates a much bigger calibration error. When the non-informative segments are included for multi-segment calibration, the calibration results may be degraded compared to calibration using only informative segments. The multi-segment calibration result is more like a weighted average of the calibration results from their component segments, thus the calibration result is more stable than single-segment calibration.

\subsubsection{\textit{Casual-walk} Sequence}
The \textit{Casual-walk} sequence is one of the common cases where the dataset is publicly available but without providing carefully calibrated extrinsic parameters. The rough extrinsic parameters are $[0,0,0]^{\top}$ in translation and identity matrix in rotation. We try to conduct multi-segment calibration on this sequence.
This sequence was collected by randomly walking around the campus and occasionally shaking the sensor vigorously, meaning that the sequence is coupled with a number of segments suitable for calibration. We perform multi-segment data calibration using only half of the sequence which is divided into 20-second segments, and the results are shown in Table~\ref{tab:segment_mit}. 
Since there are no reliable reference values for the dataset's extrinsics or ground truth trajectory, we can only evaluate the quality of the LiDAR point cloud map, which implicitly reflects the calibration accuracy. The MMEs of the maps are -2.141, -2.163, and -2.185 for the rough extrinsics, the calibrated spatial extrinsics, and the calibrated spatial-temporal extrinsics, respectively. 
The map quality obtained from the calibrated spatial-temporal extrinsics is the best, which verifies the effectiveness and accuracy of the multi-segment calibration.

\begin{table}[t]
\caption{Evaluation of calibration results using single or multi informative segments over the \textit{Casual-walk} sequence. 
}
\centering
\resizebox{\linewidth}{!}
{
    \begin{tabular}{|c|ccccccc|}
    \hline
     & \multicolumn{3}{c|}{\cellcolor[HTML]{E0EFD4}${}^I\bp_{L}$ (cm)} & \multicolumn{3}{c|}{\cellcolor[HTML]{E0EFD4}${}^I_L\bar{q}$ in Euler (deg)} & \cellcolor[HTML]{E0EFD4} \\ \cline{2-7}
    \multirow{-2}{*}{Data} & \cellcolor[HTML]{E0EFD4}$p_x$ & \cellcolor[HTML]{E0EFD4}$p_y$ & \multicolumn{1}{c|}{\cellcolor[HTML]{E0EFD4}$p_z$} & \cellcolor[HTML]{E0EFD4}roll & \cellcolor[HTML]{E0EFD4}pitch & \multicolumn{1}{c|}{\cellcolor[HTML]{E0EFD4}yaw} & \multirow{-2}{*}{\cellcolor[HTML]{E0EFD4}\begin{tabular}[c]{@{}c@{}}Time \\Offset (ms)\end{tabular}} \\ \hline
    Segment-1  & -0.9        & 0.4         & -4.7       & 178.88     & 179.83     & 178.52    & 8.3  \\ 
    Segment-2  & -0.8        & 0.7         & -2.9       & 178.92     & 179.81     & 178.53    & 8.1  \\ 
    Segment-3  & -1.6        & -0.1        & -3.2       & 178.98     & 179.90     & 178.50    & 9.1  \\ 
    Segment-4  & -0.2        & 0.0         & -1.2       & 178.91     & 179.77     & 178.72    & 8.5  \\ 
    \hline
    2 Segments & -1.0        & 0.5         & -3.8       & 178.89     & 179.81     & 178.52    & -    \\ 
    3 Segments & -1.3        & 0.4         & -3.7       & 178.91     & 179.83     & 178.52    & -    \\ 
    4 Segments & -1.1        & 0.7         & -3.3       & 178.91     & 179.83     & 178.54    & -    \\ 
	\hline
    \end{tabular}
}
\label{tab:segment_mit}
\end{table}

\begin{figure}[t]
    \begin{subfigure}{1.0\columnwidth} 
	    \centering
		\includegraphics[width=1.0\columnwidth]{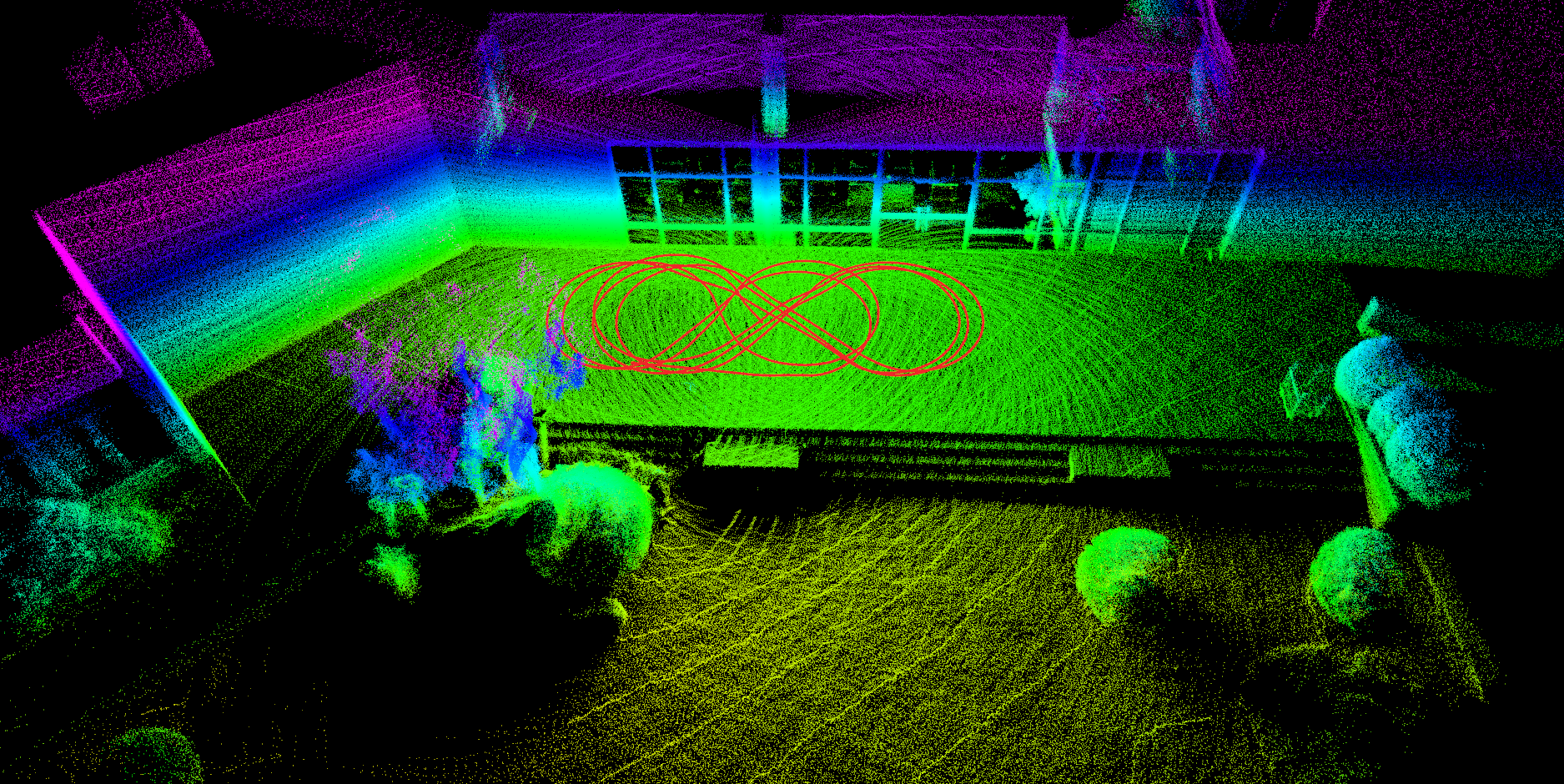}
    \end{subfigure}
	\begin{subfigure}{1.0\columnwidth} 
	    \centering
		\includegraphics[width=\columnwidth]{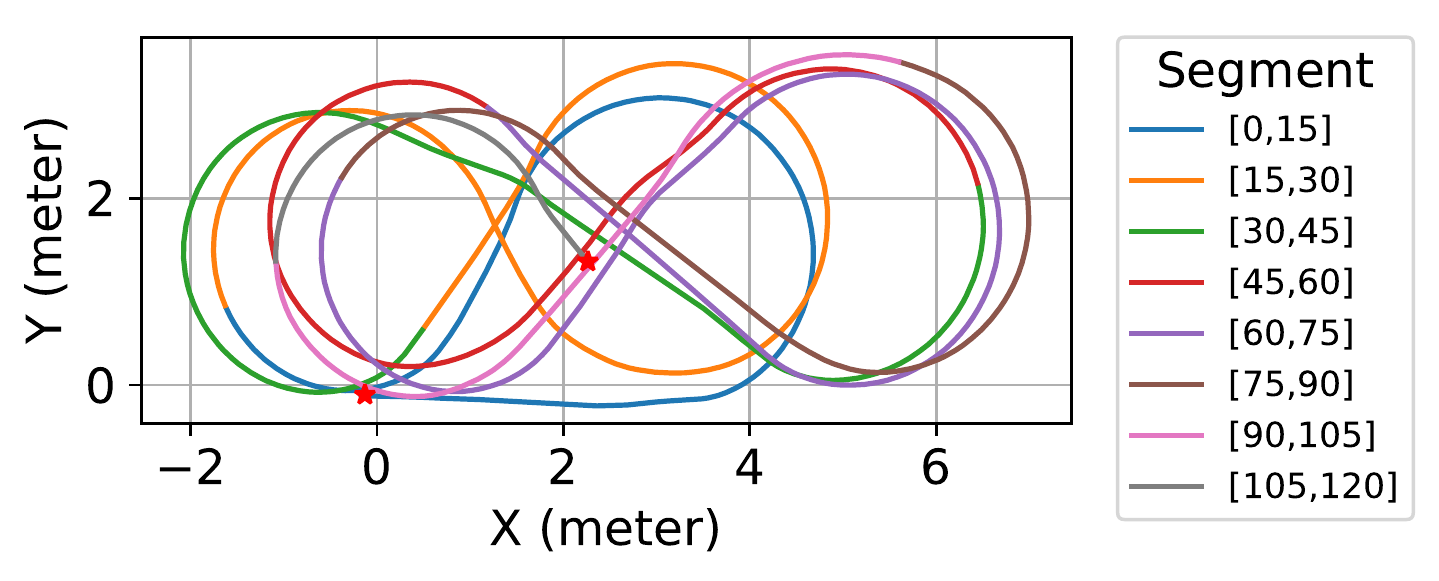}
    \end{subfigure}
    \begin{subfigure}{1.0\columnwidth} 
	    \centering
		\includegraphics[width=\columnwidth]{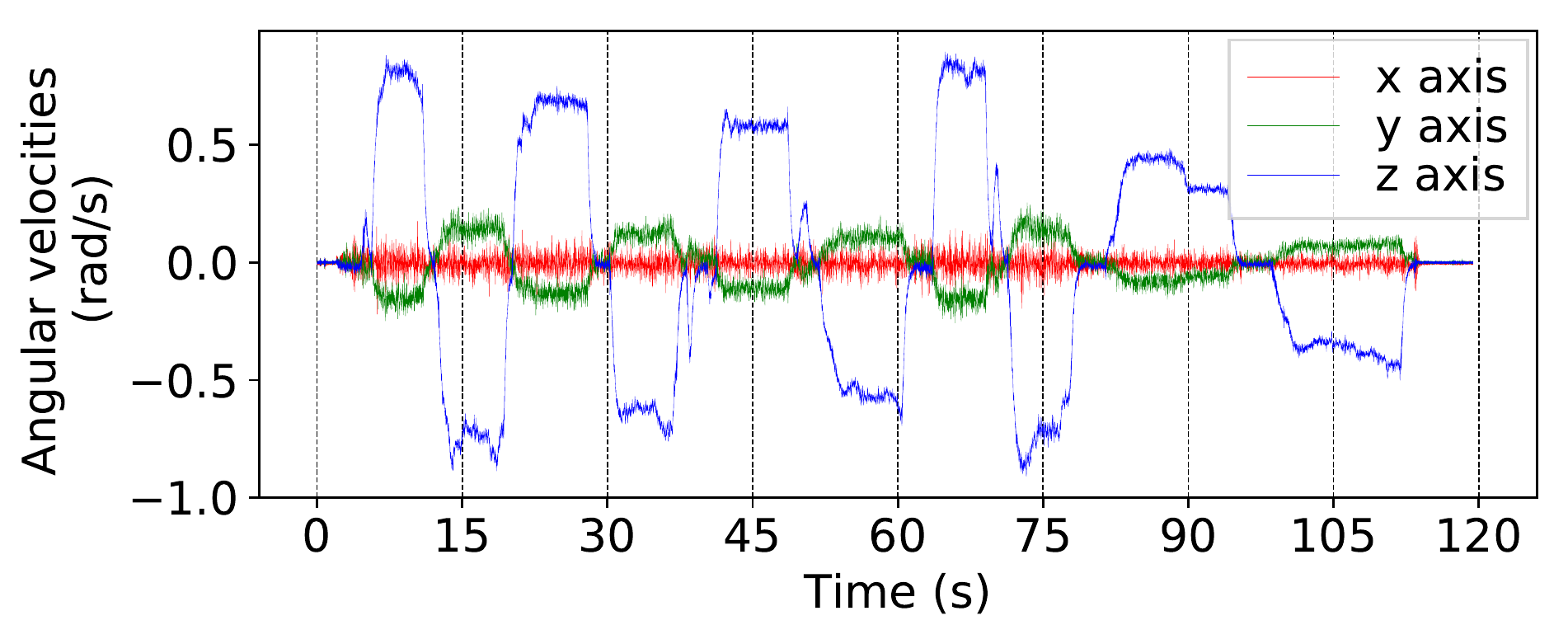}
    \end{subfigure}
	\caption{The details about the \textit{SKG-01} sequence. Top: The environment for collecting \textit{SKG-01} sequence and the actual figure-8-shape trajectory. Middle: Concrete trajectories for each 15-second segment, the red stars indicate the start and end position of the trajectory, respectively. Bottom: Corresponding three-axis angular velocity curves.}
	\label{fig:real_fig8_data}
\end{figure}

\begin{table}[t]
\centering
\begin{tabular}{ccccccc}
\toprule
\textbf{Time}          & \multicolumn{6}{c}{\textbf{Singular values in a decreasing order}}       \\
\midrule
\rowcolor[HTML]{EFEFEF} 
{[}0,15{]}    & 5545  & 4659  & 3664 & 958  & 282 & 0.010 \\
{[}15,30{]}   & 33595 & 16045 & 6844 & 930  & 724 & 0.016 \\
\rowcolor[HTML]{EFEFEF} 
{[}30,45{]}   & 27126 & 10997 & 7585 & 1915 & 678 & 0.016 \\
{[}45,60{]}   & 47269 & 25342 & 5606 & 726  & 365 & 0.009 \\
\rowcolor[HTML]{EFEFEF} 
{[}60,75{]}   & 9332  & 9186  & 3634 & 929  & 452 & 0.008 \\
{[}75,90{]}   & 60975 & 15907 & 5675 & 563  & 234 & 0.004 \\
\rowcolor[HTML]{EFEFEF} 
{[}90,105{]}  & 18542 & 7932  & 760  & 290  & 125 & 0.006 \\
{[}105,120{]} & 20997 & 10650 & 3940 & 811  & 8   & 0.012 \\ \bottomrule
\end{tabular}
\caption{The singular value of each segment in \textit{SKG-01} sequence identified by OA-LICalib. The secondary minimum singular value of the segments [90,105] and [105,120] are relatively small.
}
\label{tab:real_fig8_singular_value}
\vspace{-1em}
\end{table}

\begin{figure}[t]
	\centering
    \includegraphics[width=1.0\columnwidth]{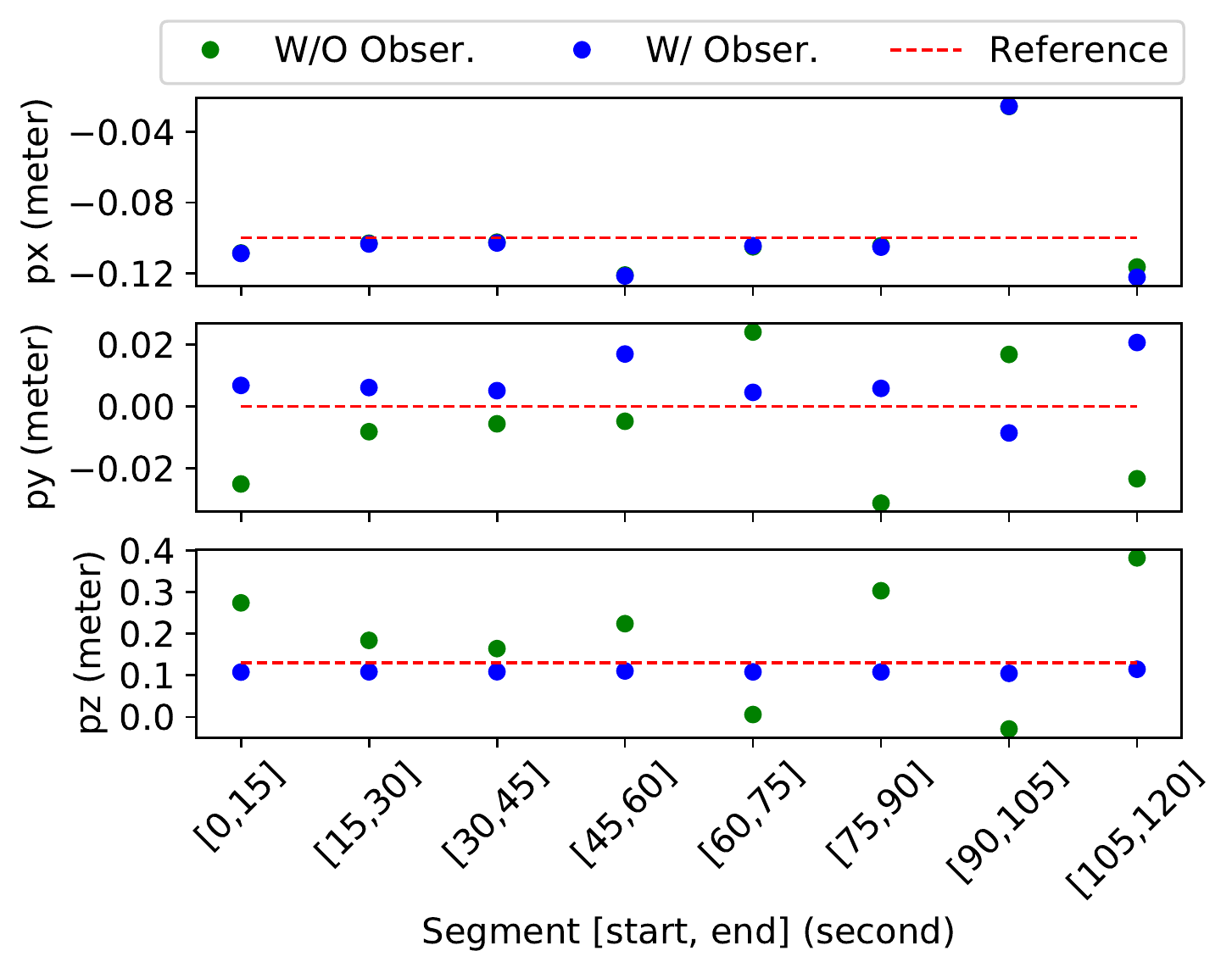}
	\caption{Comparison of the estimated ${}^I\bp_L$ under degenerate case over \textit{SKG-01} sequence: using OA-LICalib with and without observability-aware calibration. References are from the CAD sketch.}
	\label{fig:real_fig8_result}
\vspace{-1em}	
\end{figure}

\begin{figure}[t]
	\centering
    \includegraphics[width=1.0\columnwidth]{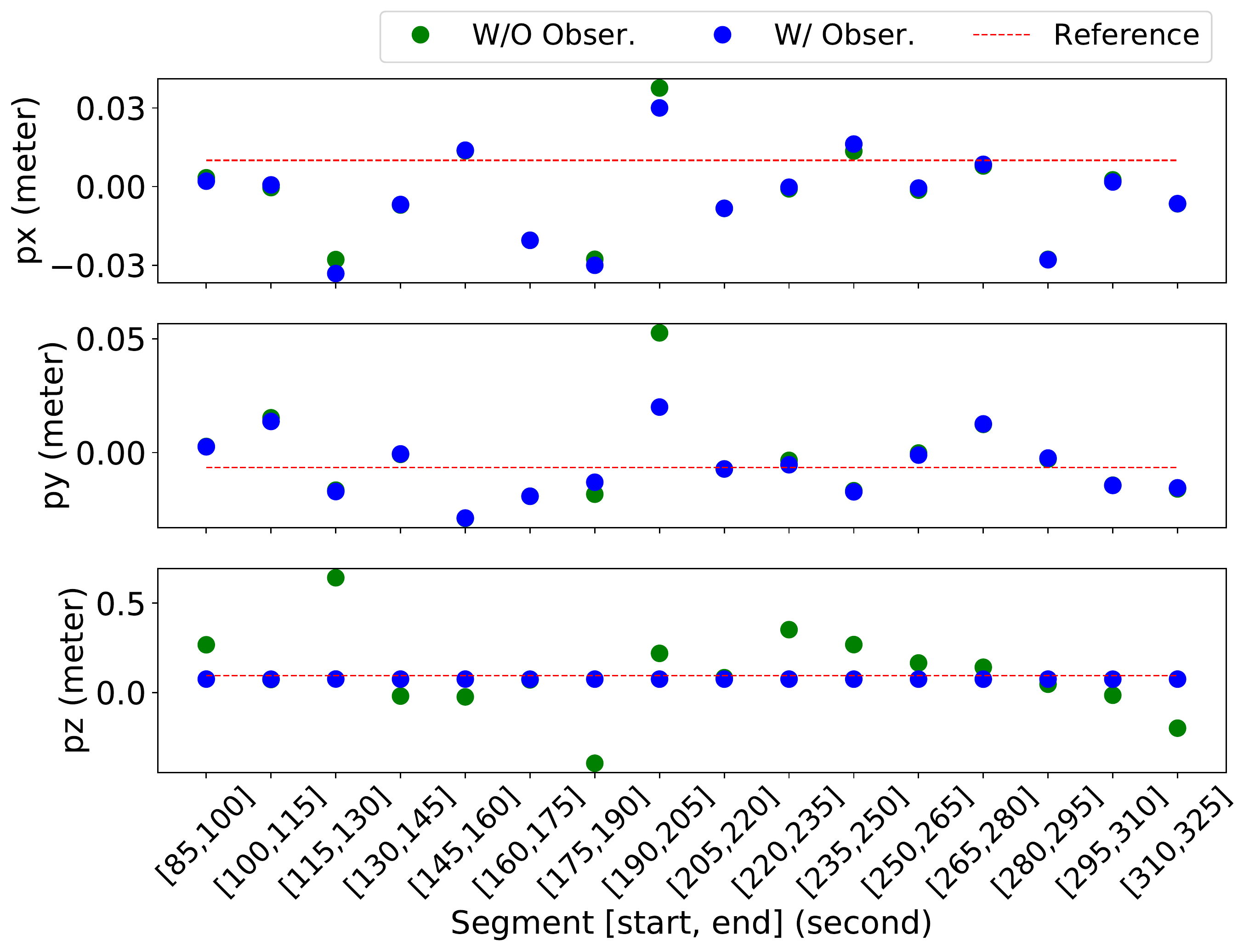}
	\caption{Comparison of the estimated ${}^I\bp_L$ under degenerate case over \textit{Basement} sequence: using OA-LICalib with and without observability awareness. 
	The beginning of the sequence is skipped due to less motion.
	}
	\label{fig:hilti_basement_calib}
\vspace{-1em}
\end{figure}

\begin{table}[t]
\caption{Comparison of the calibration results under degenerate case over \textit{SKG-01} and \textit{Basement} sequences.}
\resizebox{\linewidth}{!}{
\begin{tabular}{|c|c|ccccccc|cc|}
\hline
 & \multicolumn{1}{c|}{} & \multicolumn{1}{c|}{} & \multicolumn{3}{c|}{\cellcolor[HTML]{E0EFD4}${}^I\bp_L$ (cm)} & \multicolumn{3}{c|}{\cellcolor[HTML]{E0EFD4}${}^I_L\bar{q}$ in Euler (deg)} & \multicolumn{2}{c|}{\cellcolor[HTML]{EECC89}RMSE} \\ \cline{4-11} 
\multirow{-2}{*}{Seq.} & \multicolumn{1}{c|}{\multirow{-2}{*}{Method}} & \multicolumn{1}{c|}{\multirow{-2}{*}{Data}} & \cellcolor[HTML]{E0EFD4}$p_x$ & \cellcolor[HTML]{E0EFD4}$p_y$ & \multicolumn{1}{c|}{\cellcolor[HTML]{E0EFD4}$p_z$} & \cellcolor[HTML]{E0EFD4}roll & \cellcolor[HTML]{E0EFD4}pitch & \cellcolor[HTML]{E0EFD4}yaw & \cellcolor[HTML]{EECC89}Pos (cm) & \cellcolor[HTML]{EECC89}Rot (deg) \\ \hline
 & Reference &  & -10.00 & 0.00 & 13.00 & 0.00 & 0.00 & 90.00 &  &  \\ \cline{2-11} 
 & FAST-LIO2 & {[}0,120{]} & 4.24 & 2.09 & 11.79 & 0.59 & 0.97 & 90.77 & 8.34 & 0.79 \\
 & ILC & {[}0,120{]} & -10.50 & -0.60 & 15.35 & -0.51 & 0.13 & 91.13 & \textbf{1.43} & \textbf{0.72} \\ \cline{2-11} 
 & W/O Obser. & {[}0,15{]} & -10.85 & -2.50 & 27.39 & -0.16 & 0.29 & 90.81 & 8.45 & 0.50 \\
 & W/O Obser. & {[}15,30{]} & -10.30 & -0.82 & 18.39 & -0.12 & 0.31 & 90.64 & 3.15 & 0.42 \\
 & W/O Obser. & {[}30,45{]} & -10.24 & -0.56 & 16.41 & -0.19 & 0.28 & 91.01 & \textbf{2.00} & 0.61 \\
 & W/O Obser. & {[}45,60{]} & -12.10 & -0.48 & 22.39 & -0.10 & 0.26 & 90.56 & 5.56 & \textbf{0.36} \\ \cline{2-11} 
 & W/ Obser. & {[}0,15{]} & -10.88 & 0.68 & 10.78 & -0.16 & 0.29 & 90.80 & 1.43 & 0.50 \\
 & W/ Obser. & {[}15,30{]} & -10.34 & 0.61 & 10.84 & -0.12 & 0.31 & 90.64 & 1.31 & 0.41 \\
 & W/ Obser. & {[}30,45{]} & -10.29 & 0.51 & 10.86 & -0.19 & 0.28 & 91.01 & \textbf{1.28} & 0.61 \\
\multirow{-11}{*}{\textit{SKG-01}} & W/ Obser. & {[}45,60{]} & -12.15 & 1.69 & 11.04 & -0.10 & 0.26 & 90.56 & 1.94 & \textbf{0.36} \\ \hline \hline
 &  & \multicolumn{1}{c|}{} & \multicolumn{3}{c|}{\cellcolor[HTML]{E0EFD4}${}^I\bp_L$ (cm)} & \multicolumn{3}{c|}{\cellcolor[HTML]{E0EFD4}${}^I_L\bar{q}$ in Euler (deg)} & \multicolumn{2}{c|}{\cellcolor[HTML]{EECC89}RMSE} \\ \cline{4-11} 
\multirow{-2}{*}{Seq.} & \multirow{-2}{*}{Method} & \multicolumn{1}{c|}{\multirow{-2}{*}{Data}} & \cellcolor[HTML]{E0EFD4}$p_x$ & \cellcolor[HTML]{E0EFD4}$p_y$ & \multicolumn{1}{c|}{\cellcolor[HTML]{E0EFD4}$p_z$} & \cellcolor[HTML]{E0EFD4}roll & \cellcolor[HTML]{E0EFD4}pitch & \cellcolor[HTML]{E0EFD4}yaw & \cellcolor[HTML]{EECC89}Pos (cm) & \cellcolor[HTML]{EECC89}Rot (deg) \\ \hline
 & Reference &  & 1.00 & -0.66 & 9.47 & 0.02 & -0.37 & -0.14 &  &  \\ \cline{2-11} 
 & FAST-LIO2 & {[}85,325{]} & 0.96 & 1.53 & 7.33 & 1.22 & 0.91 & 1.51 & \textbf{1.77} & \textbf{1.39} \\
 & ILC & {[}85,325{]} & 11.05 & 10.67 & -46.97 & 4.42 & 5.38 & 9.01 & 33.74 & 6.74 \\ \cline{2-11} 
 & W/O Obser. & {[}85,100{]} & 0.34 & 0.27 & 26.72 & 0.16 & -0.34 & 0.27 & 9.98 & 0.25 \\
 & W/O Obser. & {[}100,115{]} & -0.04 & 1.53 & 7.28 & 0.31 & -0.46 & -0.29 & \textbf{1.88} & 0.19 \\
 & W/O Obser. & {[}115,130{]} & -2.78 & -1.65 & 64.22 & 0.10 & -0.41 & 1.25 & 31.69 & 0.80 \\
 & W/O Obser. & {[}130,145{]} & -0.70 & -0.08 & -1.99 & 0.19 & -0.22 & -0.12 & 6.70 & \textbf{0.13} \\ \cline{2-11} 
 & W/ Obser. & {[}85,100{]} & 0.21 & 0.25 & 7.52 & 0.16 & -0.34 & 0.26 & \textbf{1.32} & 0.24 \\
 & W/ Obser. & {[}100,115{]} & 0.06 & 1.37 & 7.47 & 0.31 & -0.45 & -0.29 & 1.73 & 0.19 \\
 & W/ Obser. & {[}115,130{]} & -3.32 & -1.71 & 7.56 & 0.10 & -0.41 & 1.30 & 2.79 & 0.83 \\
\multirow{-11}{*}{\textit{Basement}} & W/ Obser. & {[}130,145{]} & -0.68 & -0.06 & 7.51 & 0.19 & -0.22 & -0.12 & 1.53 & \textbf{0.13} \\ \hline
\end{tabular}
}
\label{tab:degenerate_seqs_compare}
\vspace{-1em}
\end{table}

\subsection{Calibration in Degenerate Case}

In practical applications, the on-vehicle situation is a common case with motion degeneration and there are high demands for accurate calibrations in this scenario. In this section, we also test the extrinsic calibration in on-vehicle conditions over the self-collected \textit{SKG-01} sequence and the publicly-available \textit{Basement} sequence. {
Regarding the initialization under degenerate cases, since the compared methods ILC, LIOM, and FAST-LIO2 fail to be initialized with the identity extrinsic rotation matrix, all the methods are initialized with the perturbed extrinsics, with rotational perturbations of 2 degrees per axis and translational perturbations of 2 cm per axis.
In general, all the methods start from the same initial extrinsic parameters for fair comparisons.
}

\subsubsection{\textit{SKG-01} Sequence}
To ensure the LiDAR sensor has sufficiently large visibility of the ground for localization, we mount the sensor suite at a slight incline downwards as shown in Fig.~\ref{fig:car}. We collect the \textit{SKG-01} sequence with a duration of 120 seconds, while the ground robot moves on a planar ground surface. 
This sequence is split into six 15-second segments, which are individually used for calibration. Figure~\ref{fig:real_fig8_data} illustrates the details of \textit{SKG-01} sequence and Fig.~\ref{fig:real_fig8_result} displays translation calibration results of OA-LICalib with and without observability awareness (Section~\ref{sec:tsvd}). Table~\ref{tab:real_fig8_singular_value} reports the singular value of each segment in \textit{SKG-01} sequence identified by OA-LICalib. For the first 6 segments, the eigen vector corresponding to the minimum singular value is almost the same that is $[0.000,\ 0.000,\ 0.000,\ 0.007,\ -0.187,\ 0.982]$. Note that the head part of the eigen vector corresponds to extrinsic rotation and the tail part corresponds to extrinsic translation, which is determined by the stack order of parameters $\delta \bx$ in \eqref{eq:Ax=b}. The identified degradation direction by the eigen vector is close to the $z$-axis, which is reasonable and coincides with the fact that a certain angular offset exists between the sensor frame and the ground plane.
As shown in Fig.~\ref{fig:real_fig8_result}, translation results in the $z$-axis of OA-LICalib without observability awareness has a large deviation from the reference value.
With the awareness of observability, OA-LICalib is able to keep the main degenerate axis, $z$-axis, as a prior and eliminate the coupled errors in the $x$-axis, resulting in more accurate calibrated extrinsic parameters. However, for segments at time intervals [90, 105] and [105,120], OA-LICalib estimates poor results which is most likely due to insufficient excitation. From the angular velocities curves of the \textit{SKG-01} sequence shown in Fig.~\ref{fig:real_fig8_data}, we can find that the magnitudes of angular velocity are relatively small for these two segments, leading to a deterioration of the observability, which is also reflected by the singular values shown in Table~\ref{tab:real_fig8_singular_value}.

\subsubsection{\textit{Basement} Sequence}
We further test in the on-vehicle \textit{Basement} sequence of the Hilti dataset.
The beginning of the \textit{Basement} is skipped due to less motion.
Fig.~\ref{fig:hilti_basement_calib} reports the calibration results. OA-LICalib with observability awareness automatically identifies the principal non-observable direction as the $z$-axis of extrinsic translation and succeeds in keeping it to the initial value. 
Besides, we notice that the errors of final calibration results in the observable $x$-axis and $y$-axis of translations are larger than those on the \textit{SKG-01} sequence. Because the Hilti dataset is not specifically collected for calibration, and the motion excitation in this sequence is insufficient, which is also reflected by the small magnitude of averaged angular velocity shown in Table~\ref{tab:real_datasets}.

Table~\ref{tab:degenerate_seqs_compare} reports the calibration results for the different methods. Again, LIOM fails to give reasonable results thus are excluded from the table.
The main degenerate direction of \textit{SKG-01} and \textit{Basement} sequences is the extrinsic translation along the $z$-axis.
Without the observability-aware module, our method, FAST-LIO2, and ILC fail to give good calibrations along the $z$-axis of extrinsic translation. With observability awareness, OA-LICalib is able to generate accurate calibration results in on-vehicle conditions.

\subsection{Remarks}
\label{sec:disscussion}

\rev{It is clear from the previous simulation and experimental results that fully-excited motions, whenever possible, are generally recommended for accurate LI calibration, which agrees with the calibration literature~\cite{yang2020online,kelly2011visual,furgale2012continuous}.
For example, 
the proposed OA-LICalib is able to provide accurate calibration results from any segment with fully-excited motion in sequence \textit{Lab} and \textit{Stairs}, as shown in Fig.~\ref{fig:6dof-position} and Fig.~\ref{fig:6dof-rotation} and Table~\ref{tab:real_full_motion_compare}. Among the sequences with occasional fully-excited motion, e.g., \textit{YQ-01}, \textit{IC\_Office} and \textit{Office\_Mitte} sequences, the automatically-selected informative segments can also generate reasonable calibration results as summarized in Table~\ref{table:segment_skg}, ~\ref{tab:hilti_multi_segment}. 
In contrast,
the OA-LICalib has larger errors on segments under degenerate planar motions, such as [90,105] of the \textit{SKG-01} sequence (see Fig.~\ref{fig:real_fig8_result}) 
and [115,130] of the \textit{Basement} sequence (see Fig.~\ref{fig:hilti_basement_calib}), which are the segments with small angular velocities (see Fig.~\ref{fig:real_fig8_data}).
}

\rev{While the fully-exited motions are recommended, they might not be feasible in practice and it thus becomes necessary to address degeneracy.
The proposed OA-LICalib with observability-awareness is able to select informative segments based on the singular values of the information matrix and tackle the degeneracy via TSVD. 
Notably, the information metric of singular values is affected by the sensory observations, motion profiles and  environments. 
Some other information metrics, for example, the trace, determinant, and maximal eigenvalue of the information matrix can also be used~\cite{schneider2019observability}. 
}

\rev{It should be noted that the proposed OA-LICalib relies on the point-to-surfel constraints, which well constrain the LI calibration problem in  structured environments. 
This implies that 
planar patches (surfels) exist in the environments and can be identified. 
From our experience, no matter it is indoors or outdoors, the OA-LICalib usually works well when there are enough man-made structures (e.g., walls, buildings, staircases) within 50 meters and the calibration data collection with fully-excited motion lasts for 10-20 seconds.
However, in some  cluttered environments without clear planar regions or some scenarios with fairly sparse structures, the proposed approach would have degraded performance. 
For example, it may fail to provide reasonable results on some outdoor sequences of the Hilti dataset~\cite{helmberger2021hilti}, in which the scenarios are nearly empty without clear human-made structures.
To address this issue, resulting from insufficient point-to-surfel constraints from LiDAR data, some ideas from advanced point cloud registration methods~\cite{huang2021comprehensive} could be leveraged, such as Sdrsac~\cite{le2019sdrsac}, Teaser++~\cite{yang2019polynomial, yang2020teaser}, OPRANSAC~\cite{li2021point}, CVO~\cite{zhang2021new}, and 
deep-learning-based methods~\cite{huang2021predator}.
}

\section{Conclusions and Future Work}
\label{sec:conclusion}

\rev{
In this paper, we have developed an observability-aware targetless LiDAR-IMU calibration method, termed OA-LICalib, within the continuous-time batch optimization framework. 
The proposed OA-LICalib calibrates not only the spatial-temporal extrinsic parameters but also the intrinsic parameters of both IMU and LiDAR sensors,
while enforcing observability constraints during update to address possible degeneracy.
Specifically, two observability-aware strategies are employed: (i) the informative data segment selection,  and (ii) the observability-aware update in  back-end optimization.
The former selects the most informative data segments automatically for calibration among the long-session data sequence, which lowers the computational consumption for calibration as well as helps non-expert end users. 
The latter addresses the degenerate motions via truncated SVD (TSVD) that is used  to update only the observable directions of the sensing parameters.
The proposed method has been extensively validated on both Monte-Carlo simulations and real-world experiments,
showing that the proposed OA-LICalib is able to provide accurate spatial-temporal extrinsic and intrinsic calibration with high repeatability, even under certain degenerate cases.
}
For future work, the active calibration guiding users to collect informative data deserves to be investigated.

\section*{Acknowledgments}
This work is supported by a Grant from the National Natural Science Foundation of China (No. U21A20484) and Huang was partially supported by the University of Delaware
College of Engineering.

{
	\vspace{0.06cm}
    \def\bibfont{\footnotesize}
	\printbibliography
}

\end{document}